\documentclass[journal]{IEEEtran}

\usepackage[pdftex]{graphicx}
\usepackage{amsmath,color,bm,amssymb,mathrsfs,subfigure,booktabs,cite,url,mathrsfs}

\graphicspath{{fig/}}

\begin{document}

\setlength{\abovedisplayskip}{4pt}
\setlength{\belowdisplayskip}{4pt}
\setlength{\arraycolsep}{0.1em}
\setlength{\aboverulesep}{1pt}
\setlength{\belowrulesep}{1.5pt}

\title{
Note Value Recognition for Piano Transcription Using Markov Random Fields
}
\author{Eita Nakamura,~\IEEEmembership{Member,~IEEE,}
        Kazuyoshi Yoshii,~\IEEEmembership{Member,~IEEE,}
        Simon Dixon
\thanks{E.~Nakamura is with the Graduate School of Informatics, Kyoto University, Kyoto 606-8501, Japan. He is supported by the JSPS research fellowship (PD). Electric address: {\tt enakamura@sap.ist.i.kyoto-u.ac.jp}. This work was done while he was a visiting researcher at Queen Mary University of London.}%
\thanks{K.~Yoshii is with the Graduate School of Informatics, Kyoto University, Kyoto 606-8501, Japan and with AIP, RIKEN, Tokyo 103-0027, Japan.}
\thanks{S.~Dixon is with the School of Electronic Engineering and Computer Science, Queen Mary University of London, London E1 4NS, UK.}
\thanks{Manuscript received XX, YY; revised XX, YY.}}

\markboth{Journal of \LaTeX\ Class Files,~Vol.~XX, No.~YY, ZZZZ}%
{Nakamura \MakeLowercase{\textit{et al.}}: Note Value Recognition for Piano Transcription}

%


\maketitle

\begin{abstract}
This paper presents a statistical method for use in music transcription that can estimate score times of note onsets and offsets from polyphonic MIDI performance signals. Because performed note durations can deviate largely from score-indicated values, previous methods had the problem of not being able to accurately estimate offset score times (or note values) and thus could only output incomplete musical scores. Based on observations that the pitch context and onset score times are influential on the configuration of note values, we construct a context-tree model that provides prior distributions of note values using these features and combine it with a performance model in the framework of Markov random fields. Evaluation results show that our method reduces the average error rate by around 40 percent compared to existing/simple methods. We also confirmed that, in our model, the score model plays a more important role than the performance model, and it automatically captures the voice structure by unsupervised learning.
\end{abstract}


%
\IEEEpeerreviewmaketitle

\section{Introduction}\label{sec:Intro}

Music transcription is one of the most fundamental and challenging problems in music information processing \cite{Klapuri2006,Benetos2013}.
This problem, which involves conversion of audio signals into symbolic musical scores, can be divided into two subproblems, pitch analysis and rhythm transcription, which are often studied separately.
Pitch analysis aims to convert the audio signals into the form of a piano roll, which can be represented as a MIDI signal, and multi-pitch analysis methods for polyphonic music have been extensively studied \cite{Vincent2010,OHanlon2014,Yoshii2015,Sigtia2016}.
Rhythm transcription, on the other hand, aims to convert a MIDI signal into a musical score by locating note onsets and offsets in musical time ({\it score time}) \cite{LonguetHiggins1987,Desain1989,Raphael2002,Takeda2002,Hamanaka2003,Cemgil2003,Kapanci2005,Temperley2009,Tsuchiya2013,Nakamura2017}.
In order to track time-varying tempo, beat tracking is employed to locate beat positions in music audio signals \cite{Dixon2000,Dixon2001,Peeters2011,Krebs2015,Durand2015}.

Although most studies on rhythm transcription and beat tracking have focused on estimating onset score times, to obtain complete musical scores it is necessary to locate note offsets, or equivalently, identify {\it note values} defined as the difference between onset and offset score times.
The configuration of note values is especially important to describe the acoustic and interpretative nature of polyphonic music where there are multiple voices and the overlapping of notes produces different harmonies.
Note value recognition has been addressed only in a few studies \cite{Takeda2002,Temperley2009} and the results of this study reveal that it is a non-trivial problem.

The difficulty of the problem arises from the fact that observed note durations in performances deviate largely from the score-indicated lengths so that the use of a prior (language) model for musical scores is crucial.
Because of its structure with overlapping multiple streams (voices), construction of a language model for polyphonic music is challenging and gathers increasing attention recently \cite{Temperley2009,Kameoka2012,Raczynski2013,Sigtia2016,Nakamura2017}.
In particular, building a model at the symbolic level of musical notes (as opposed to the frame level of audio processing) that properly describes the multiple-voice structure while retaining computational tractability is a remaining problem.

\begin{figure}[t]
\begin{center}
\includegraphics[clip,width=0.9\columnwidth]{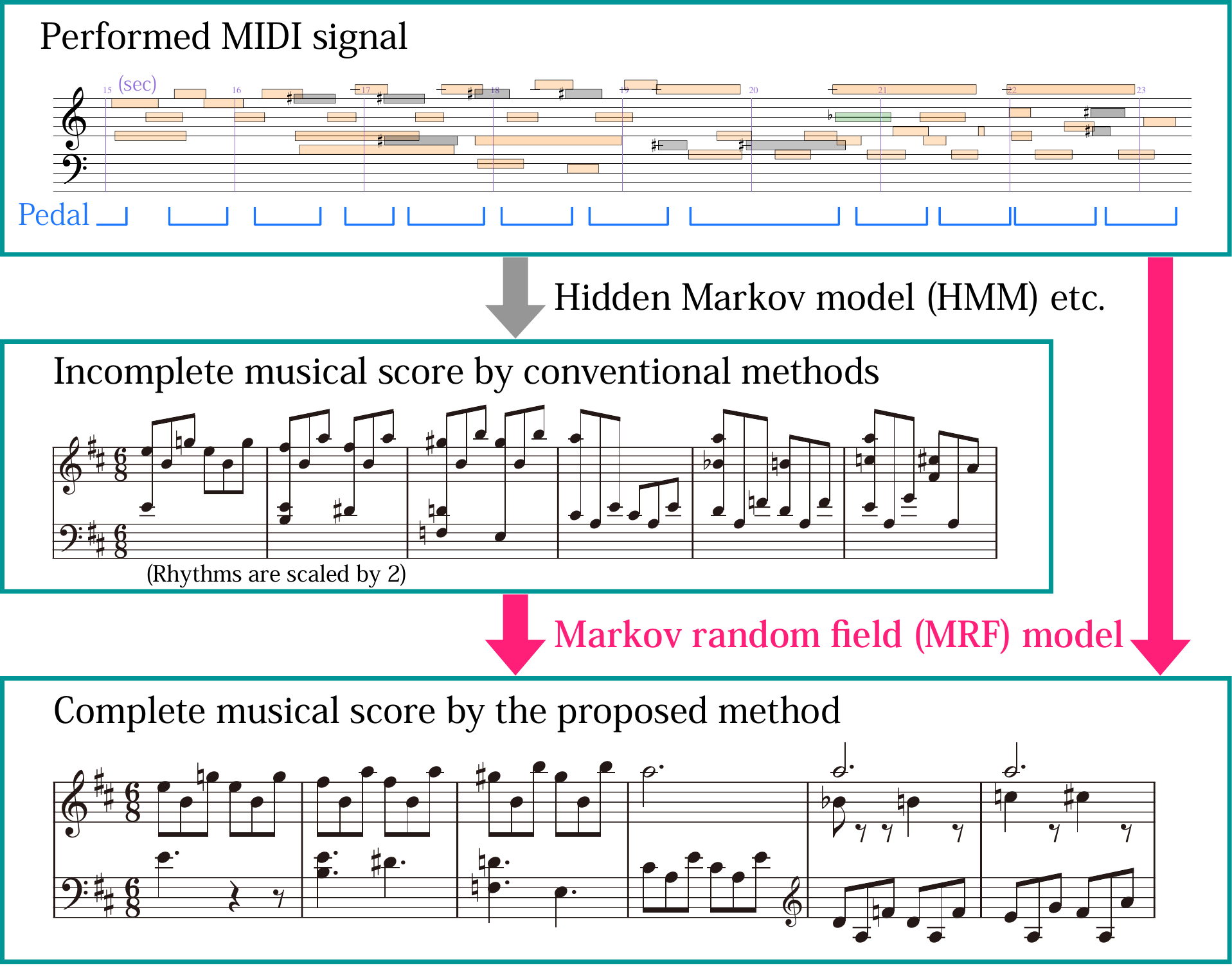}
\end{center}
\vspace{-2mm}
\caption{An outcome obtained by our method (Mozart: Piano Sonata K576). While previous rhythm transcription methods could only estimate onset score times accurately from MIDI performances, our method can also estimate offset score times, providing a complete representation of polyphonic musical scores.}
\label{fig:Overview}
\vspace{-3mm}
\end{figure}
The purpose of this paper is to investigate the problem of note value recognition using a statistical approach (Fig.~\ref{fig:Overview}).
We formulate the problem as a post-processing step of estimating offset score times given onset score times obtained by rhythm transcription methods for note onsets.
Firstly, we present results of statistical analyses and point out that the information of onset score times and the pitch context together with interdependence between note values provide clues for model construction.
Secondly, we propose a Markov random field model that integrates a prior model for musical scores and a performance model that relates note values and actual durations (Sec.~\ref{sec:Method}).
To determine an optimal set of contexts/features for the score model from data, we develop a statistical learning method based on context-tree clustering \cite{ContextTreeClustering,Takaki2014,Shinoda2000}, which is an adaptation of statistical decision tree analysis.
Finally, results of systematic evaluations of the proposed method and baseline methods are presented (Sec.~\ref{sec:Evaluation}).

The contributions of this study are as follows.
We formulate a statistical learning method to construct a highly predictive prior model for note values and quantitatively demonstrate its importance for the first time.
The discussions cover simple methods and more sophisticated machine learning techniques and the evaluation results can serve as a reference for the state-of-the-art.
Our problem is formulated in a general setting following previous studies on rhythm transcription and the method is applicable to a wide range of existing methods of onset rhythm transcription.
Results of statistical analyses and learning in Secs.~\ref{sec:Observation} and \ref{sec:Method} can also serve as a useful guide for research using other approaches such as rule-based methods and neural networks.
Lastly, source code of our algorithms and evaluation tools is available from the accompanying web page \cite{Webpage} to facilitate future comparisons and applications.

\section{Related Work}\label{sec:RelatedWork}

Before beginning the main discussion, let us review previous studies related to this paper.

There have been many studies on converting MIDI performance signals into a form of musical score.
Older studies \cite{LonguetHiggins1987,Desain1989} used rule-based methods and networks in attempts to model the process of human perception of musical rhythm.
Since around 2000, various statistical models have been proposed to combine the statistical nature of note sequences in musical scores and that of temporal fluctuations in music performance.
The most popular approach is to use hidden Markov models (HMMs) \cite{Raphael2002,Takeda2002,Hamanaka2003,Cemgil2003,Nakamura2017}.
The score is described either as a Markov process on beat positions (metrical Markov model) \cite{Raphael2002,Hamanaka2003,Cemgil2003} or a Markov model of notes (note Markov model) \cite{Takeda2002}, and the performance model is often constructed as a state-space model with latent variables describing locally defined tempos.
Recently a merged-output HMM incorporating the multiple-voice structure has been proposed \cite{Nakamura2017}.
Temperley \cite{Temperley2009} proposed a score model similar to the metrical Markov model in which the hierarchical metrical structure is explicitly described.
There are also studies that investigated probabilistic context-free grammar models \cite{Tsuchiya2013}.

A recent study \cite{Nakamura2017} reported results of systematic evaluation of (onset) rhythm transcription methods.
Two data sets, polyrhythmic data and non-polyrhythmic data, were used and it was shown that HMM-based methods generally performed better than others and the merged-output HMM was most effective for polyrhythmic data.
In addition to the accuracy of recognising onset beat positions, the metrical HMM has the advantage of being able to estimate metrical structure, i.e.\ the metre (duple or triple) and bar (or down beat) positions, and to avoid grammatically incorrect score representations that appeared in other HMMs.

As mentioned above, there have been only a few studies that discussed the recognition of note values in addition to onset score times.
Takeda et al.\ \cite{Takeda2002} applied a similar method of estimating onset score times to estimating note values of monophonic performances and reported that the recognition accuracy dropped from 97.3\% to 59.7\% if rests are included.
Temperley's Melisma Analyzer \cite{Temperley2009}, based on a statistical model, outputs estimated onset and offset beat positions together with voice information for polyphonic music.
There, offset score times are chosen from one of the following tactus beats according to some probabilities, or chosen as the onset position of the next note of the same voice.
The recognition accuracy of note values has not been reported.

\section{Preliminary Observations and Analyses}\label{sec:Observation}

We explain here basic facts about the structure of polyphonic piano scores and discuss how it is important and non-trivial to recognise note values for such music based on observations and statistical analyses.
This provides motivations for the architecture of our model.
Some terminology and notions used in this paper are also introduced.
We consider the music style of the common practice period and similar music styles such as popular and jazz music in this paper.

\subsection{Structure of Polyphonic Musical Scores}\label{sec:ScoreStructure}

%
\begin{figure}[t]
\begin{center}
\includegraphics[clip,width=0.85\columnwidth]{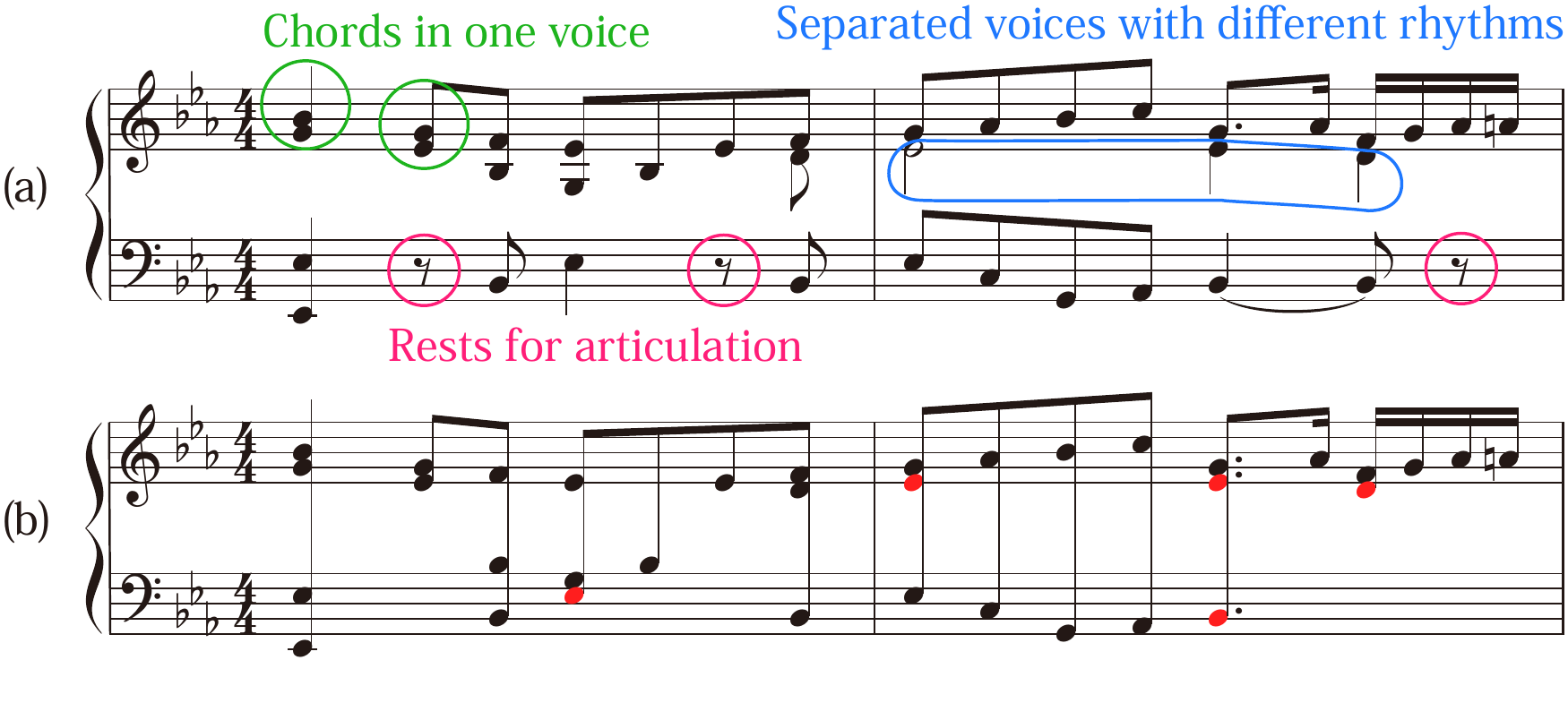}
\end{center}
\vspace{-3mm}
\caption{Example of (a) a polyphonic piano score (Mozart: Sonata KV570) and (b) a reduced score represented with one voice. Notes that have different note values in the two representations are indicated with red note heads.}
\label{fig:ExampleScore}
\vspace{-3mm}
\end{figure}
To discuss recognition of note values in polyphonic piano music, we first explain the structure of polyphonic scores.
The left-hand and right-hand parts are usually written in separate staffs and each staff can contain several {\it voices}\footnote{Our ``voice'' corresponds to the voice information defined in music notation file formats such as MusicXML and Finale file format.}, or streams of notes (Fig.~\ref{fig:ExampleScore}(a)).
In piano scores, each voice can contain chords and the number of voices can vary locally.
Hereafter we use the word {\it chords} to indicate those within one voice.
Except for rare cases of partial ties in chords, notes in a chord must have simultaneous onset and offset score times.
This means that the offset score time of a note must be equal to or earlier than the onset score time of the next note/chord of the same voice.
In the latter case, the note is followed by a rest.
Such rests are rare \cite{Temperley2009} and thus the configuration of note values and the voice structure are inter-related.

The importance of voice structure in the description of note values can also be understood by comparing a polyphonic score with a reduced score obtained by putting all notes with simultaneous onsets into a chord and forming one `big voice' without any rests as in Fig.~\ref{fig:ExampleScore}(b).
Since these two scores are the same in terms of onset score times, the differences are only in offset score times.
One can see that appropriate voice structure is necessary to recover correct note values from the reduced score.
It can also be confirmed that note values are influential to realise the expected acoustic effect of polyphonic music.
Because one can automatically obtain the reduced score given the onset score times, recovering the polyphonic score as in Fig.~\ref{fig:ExampleScore}(a) from the reduced score as in Fig.~\ref{fig:ExampleScore}(b) is exactly the aim of note value recognition.

\subsection{Distribution of Durations in Music Performances}\label{sec:DurationDistribution}

A natural approach to recover note values from MIDI performances is finding those note values that best fit the actual note durations in the performances.
In this paper, {\it duration} always means the time length measured in physical time, and a score-written note length is called a note value.
To relate durations to note values, one needs the (local) tempo that provides the conversion ratio.
Although estimating tempos from MIDI performances is a nontrivial problem (see Sec.~\ref{sec:Method}), let us suppose here they are given, for simplicity.
Given a local tempo and a note value, one can calculate an expected duration, and conversely, one can estimate a note value given a local tempo and actual duration.

\begin{figure}[t]
\begin{center}
\subfigure[]
{\includegraphics[clip,width=0.49\columnwidth]{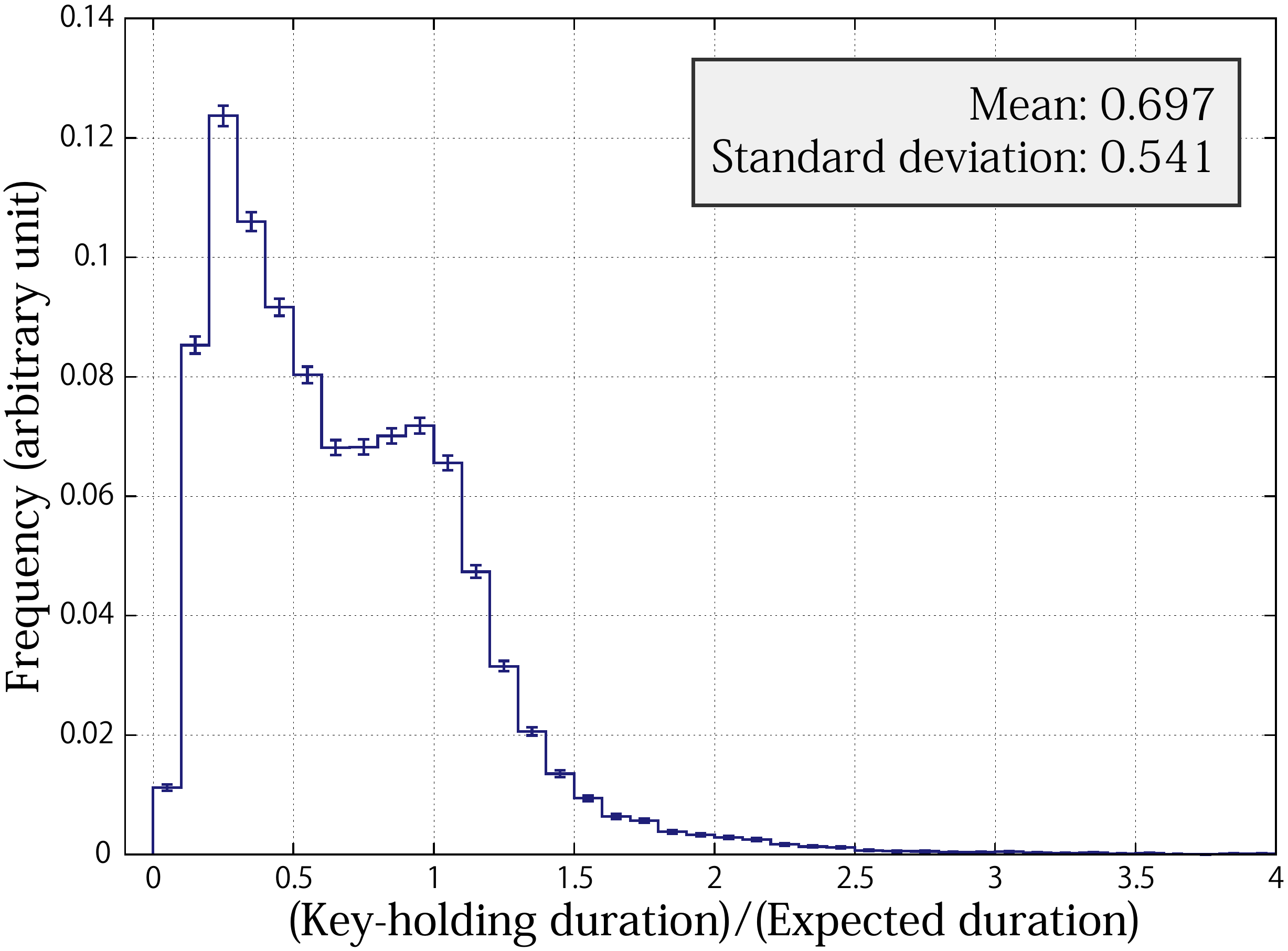}}
\subfigure[]
{\includegraphics[clip,width=0.49\columnwidth]{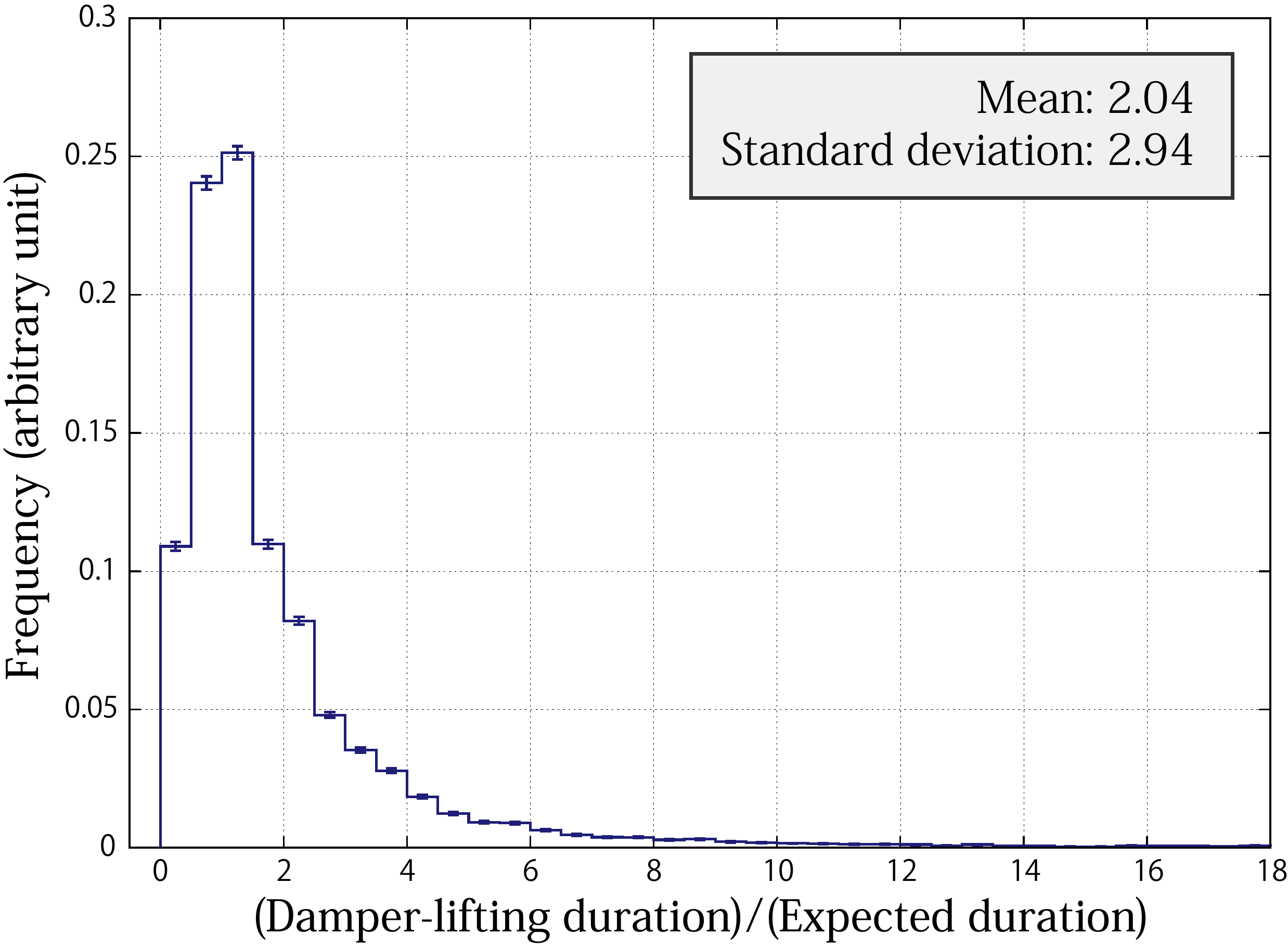}}
\vspace{-5mm}
\end{center}
\caption{Distributions of the ratios of actual duration, (a) key-holding durations and (b) damper-lifting durations, to the expected duration.}
\label{fig:Distribution}
\vspace{-4mm}
\end{figure}
Fig.~\ref{fig:Distribution} shows distributions of the ratios of actual durations in performances and the durations expected from note values and tempos estimated from onset times (used performance data is described in Sec.~\ref{sec:Learning}).
Because information of key-press and key-release times for each note and pedal movements can be obtained from MIDI signals, one can define the following two durations.
The {\it key-holding duration} is the time interval between key-press and key-release times and the {\it damper-lifting duration} is obtained by extending the offset time as long as the sustain/sostenuto pedal is held.
As can be seen from the figure, both distributions have large variances and thus precise prediction of note values is impossible by using only the observed values.
As mentioned previously \cite{Cemgil2003,Temperley2009}, this makes note value recognition a difficult problem and it has often been avoided in previous studies.
Additionally, due to the large deviations of durations, most tempo estimation methods use only onset time information.

A similar situation happens in speech recognition where the presence of acoustic variations and noise makes it difficult to extract symbolic text information by pure feature extraction.
Similarly to using a prior language model, which was the key to improve the accuracy of speech recognition \cite{Levinson1983}, a prior model for musical scores ({\it score model}) would be a key to solving our problem, which we seek in this paper.

\subsection{Hints for Constructing a Score Model}\label{sec:HintsForScoreModel}

\begin{figure}[t]
\begin{center}
\includegraphics[clip,width=1.\columnwidth]{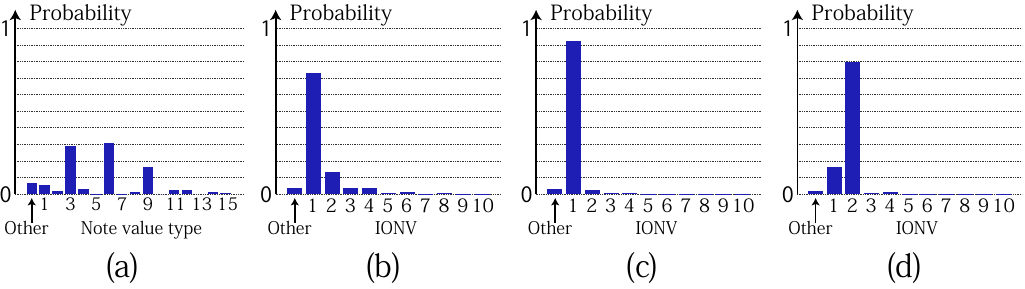}
\end{center}
\vspace{-3mm}
\caption{Distributions of note values. In (a), note values are categorised into 15 types in Eq.~(\ref{eq:NoteValues}) and another type including all others; in (b)(c)(d), they are categorised into the first ten IONVs and others. Samples in (c)(d) were selected by conditions on the pitch context described in the text.}
\label{fig:DistrIONVClass}
\vspace{-3mm}
\end{figure}

The simplest score model for note value recognition would be a discrete probability distribution over a set of note values.
For example, one can consider the following 15 types of note values (e.g.\ $1/2=\text{half note}$, $3/16=\text{dotted eighth note}$, etc.):
\begin{equation}
\big\{\tfrac{1}{32},\tfrac{1}{48},\tfrac{1}{16},\tfrac{1}{24},\tfrac{3}{32},\tfrac18,\tfrac{1}{12},\tfrac{3}{16},\tfrac14,\tfrac16,\tfrac38,\tfrac12,\tfrac13,\tfrac34,1\big\}.
\label{eq:NoteValues}
\end{equation}
The distribution taken from a score data set (see Sec.~\ref{sec:Learning}) is shown in Fig.~\ref{fig:DistrIONVClass}(a).
Although the distribution has clear tendencies, it is not sufficiently sharp to compensate the large variance of the duration distributions.
We will confirm that this simple model yields a poor recognition accuracy in Sec.~\ref{sec:Comparisons}.

\begin{figure}[t]
\begin{center}
\includegraphics[clip,width=0.65\columnwidth]{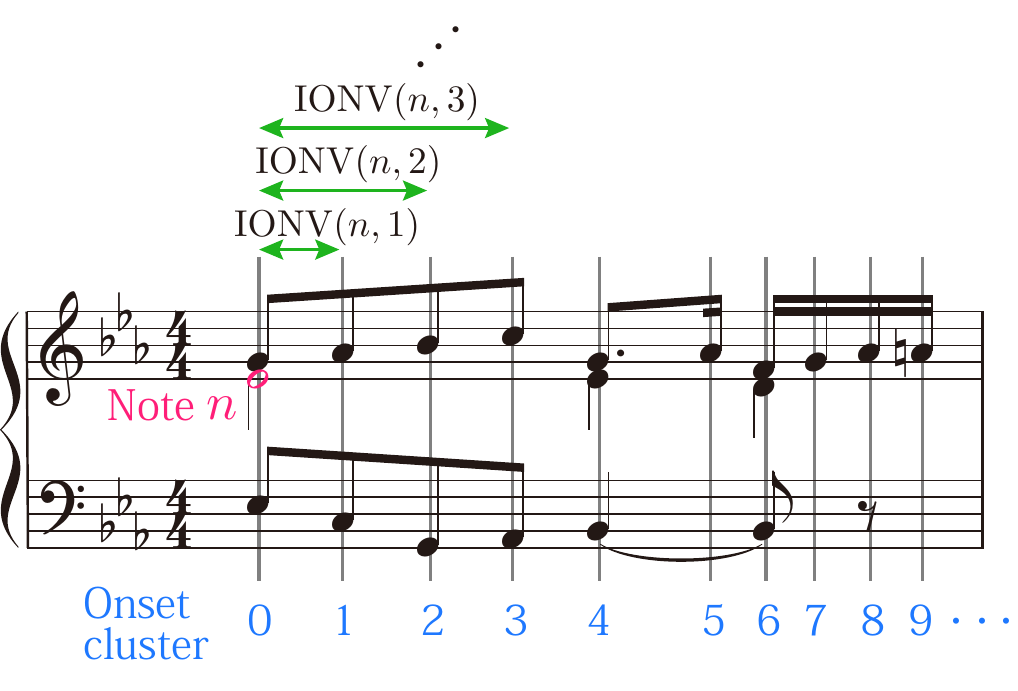}
\vspace{-4mm}
\end{center}
\caption{Onset clusters and inter-onset note values (IONVs).}
\label{fig:IONV}
\vspace{-4mm}
\end{figure}
Hints for constructing a score model can be obtained by again observing the example in Fig.~\ref{fig:ExampleScore}.
It is observed that most notes in the reduced score have the same note values as in the original score, and even when they do not, the offset score times tend to correspond with one of the onset score times of following notes.
To explain this more precisely in a statistical way, we define an {\it onset cluster} as the set of all notes with simultaneous onsets in the score and {\it inter-onset note values} ({\it IONVs}) as the intervals between onset score times of succeeding onset clusters (Fig.~\ref{fig:IONV}).
As in the figure, for later convenience, we define IONVs for each note, even though they are same for all notes in an onset cluster.
If one counts frequencies that each note value matches one of the first ten IONVs (or none of them), the result is as shown in Fig.~\ref{fig:DistrIONVClass}(b).
We see that the distribution has lower entropy than that in Fig.~\ref{fig:DistrIONVClass}(a) and the probability that note values would be different from any of the first ten IONVs is small (3.50\% in our data).
This suggests that a more efficient search space for note values can be obtained by using the onset score time information.

Even more predictive distributions of note values can be obtained by using the pitch information.
This is because neighbouring notes (either horizontally or vertically) in a voice tend to have close pitches, as discussed in studies on voice separation \cite{Cambouropoulos2008,McLeod2016,HandSeparation}.
For example, if we select notes that have a note within five semitones in the next onset cluster, the distribution of note values in the space of IONVs becomes as in Fig.~\ref{fig:DistrIONVClass}(c), reflecting the fact that inserted rests are rare.
On the other hand, if we impose a condition of having a note with five semitones in the second next onset cluster but not having any notes within 14 semitones in the next cluster, then the distribution becomes as in Fig.~\ref{fig:DistrIONVClass}(d), which reflects the fact that this condition implies that the note has an adjacent note in the same voice in the second next onset cluster.
These results suggest on one side that pitch information together with onset score time information can provide distributions of note values with more predictive ability and on the other side that those distributions are highly dependent on the pitch context.

Although so far we have considered note values as independent distributions, their interdependence can also provide clues in estimating note values.
One such interdependence can be inferred from the logical constraint of voice structure described in Sec.~\ref{sec:ScoreStructure}.
As chordal notes have the same note values and they also tend to have close pitches, notes with simultaneous onset score times and close pitches tend to have identical note values.
This is another case where pitch information has influence on the distribution of note values.

\subsection{Summary of the Section}

Here we summarise the findings in this section:
\begin{itemize}

\item The voice structure and the configuration of note values are inter-related and the logical constraints for musical scores induce interdependence between note values.

\item Performed durations contain large deviations from those implied by the score and a score model is crucial to accurately estimate note values from performance signals.

\item Information about onset score times provides an efficient search space for note values through the use of IONVs. In particular, the probability that a note value falls into one of the first ten IONVs is quite high.

\item The distribution of note values is highly dependent on the pitch context, which would be useful for improving their predictability.

\end{itemize}
In the rest of this paper, we construct a computational model to incorporate these findings and examine by numerical experiments how they quantitatively influence the accuracy of note value recognition.

\section{Proposed Method}\label{sec:Method}

\subsection{Problem Statement}\label{sec:ProblemStatement}

For rhythm transcription, the input is a MIDI performance signal, represented as a sequence of pitches, onset times and offset times $(p_n,t_n,t^{\rm off}_n,\bar{t}^{\rm off}_n)_{n=1}^N$ where $n$ is an index for musical notes and $N$ is the number of notes.
As explained in Sec.~\ref{sec:DurationDistribution}, we can define two offset times for each note, the key-release time and damper-drop time, denoted by $t^{\rm off}_n$ and $\bar{t}^{\rm off}_n$.
The corresponding key-holding and damper-lifting duration will be denoted by $d_n=t^{\rm off}_n-t_n$ and $\bar{d}_n=\bar{t}^{\rm off}_n-t_n$.
The aim is to recognise the score times of the note onsets and offsets, which are denoted by $(\tau_n,\tau^{\rm off}_n)_{n=1}^N$.
In general, $\tau_n$ and $\tau^{\rm off}_n$ take values in the set of rational numbers in units of a beat unit, say, the whole-note length.
For example, $\tau_1=0$ and $\tau^{\rm off}_1=1/4$ means that the first note is at the beginning of the score and has a quarter-note length.
We use the following notations for sequences: $\bm d=(d_n)_{n=1}^N$, $\bm\tau^{\rm off}=(\tau^{\rm off}_n)_{n=1}^N$, etc.
We call the difference $r_n=\tau^{\rm off}_n-\tau_n$ the {\it note value}.
Frequently used mathematical symbols are listed in Table \ref{tab:MathSymbols}.

\begin{table}[t]
\begin{center}
{
\begin{tabular}{ll}\toprule
Variable & Notation\\
\midrule
Index for note & $n$\\
Pitch & $p_n$\\
Onset time & $t_n$\\
Key-release [Damper-drop] (offset) time & $t^{\rm off}_n$ [$\bar{t}^{\rm off}_n$]\\
Key-holding [Damper-lifting] duration & $d_n$ [$\bar{d}_n$]\\
Onset [offset] score time & $\tau_n$ [$\tau^{\rm off}_n$]\\
Note value & $r_n$\\
Local tempo & $v_n$\\
Sequence of variables & $\bm p=(p_n)_{n=1}^N$ etc.\\
\bottomrule
\end{tabular}
}
\end{center}
\caption{List of Frequently Used Mathematical Symbols.}
\label{tab:MathSymbols}
\vspace{-7mm}
\end{table}
In this paper, we consider the situation that the onset score times $\bm\tau$ are given as estimations from conventional onset rhythm transcription algorithms.
In addition, we assume that a local tempo $v_n$, which gives a smoothed ratio of the time interval and score time interval at each note $n$, is given.
Local tempos $\bm v=(v_n)_{n=1}^N$ can be obtained from the sequences $\bm t$ and $\bm\tau$ by applying some smoothing methods such as Kalman smoothing and local averaging, and typically they can be obtained as outputs of onset rhythm transcription algorithms.

In summary, we set up the problem of note value recognition as estimating the sequence $\bm\tau^{\rm off}$ (or $\bm r$) with inputs $\bm p,\bm d,\bar{\bm d},\bm\tau$ and $\bm v$.
For concreteness, in this paper, we mainly use as $\bm\tau$ and $\bm v$ the outputs from a method based on a metrical HMM (Sec.~\ref{sec:OnsetPositionAndTempo}), but our method is applicable as a post-processing step for any rhythm transcription method that outputs $\bm\tau$.

\subsection{Estimation of Onset Score Times and Local Tempos}\label{sec:OnsetPositionAndTempo}

To estimate onset score times $\bm\tau$ and local tempos $\bm v$ from a MIDI performance $(\bm p,\bm t,\bm t^{\rm off},\bar{\bm t}^{\rm off})$, we use a metrical HMM \cite{Raphael2002}, which is one of the most accurate onset rhythm transcription methods (Sec.~\ref{sec:RelatedWork}).
Here we briefly review the model.

In the metrical HMM, the probability $P(\bm\tau)$ of the score is generated from a Markov process on periodically defined beat positions denoted by $(s_n)_{n=1}^N$ with $s_n\in\{1,\ldots,G\}$ ($G$ is a period of beats such as a bar).
The sequence $\bm s$ is generated with the initial and transition probabilities as
\begin{equation}
P(\bm s)=P(s_1)\prod_{n=2}^NP(s_n|s_{n-1}).
\label{eq:MetricalMM}
\end{equation}
We interpret $s_n$ as $\tau_n$ modulo $G$, or more explicitly, we obtain $\bm\tau$ incrementally as follows:
\begin{align}
\tau_1&=s_1,
\\
\tau_{n+1}&=\tau_n+\begin{cases}
s_{n+1}-s_n, &{\rm if}~s_{n+1}>s_n;\\
G+s_{n+1}-s_n, &{\rm if}~s_{n+1}\leq s_n.
\end{cases}
\end{align}
That is, if $s_{n+1}\leq s_n$, we interpret that $s_{n+1}$ indicates the beat position in the next bar.
With this understood, $P(\bm\tau)$ is equivalent to $P(\bm s)$ as long as $r_n\leq G$ for all $n$.
An extension is possible to allow note onset intervals larger than $G$ \cite{NakamuraEUSIPCO2016}.

In constructing the performance model, local tempo variables $\bm v$ are introduced to describe the indeterminacy and temporal variations of tempos.
The probability $P(\bm t,\bm v|\bm\tau)$ is decomposed as $P(\bm t|\bm\tau,\bm v)P(\bm v)$ and each factor is described with the following Gaussian Markov process:
\begin{align}
&P(v_{n}|v_{n-1})={\sf N}(v_{n};v_{n-1},\sigma_v^2),
\\
&P(t_{n+1}|t_n,\tau_{n+1},\tau_n,v_n)
\notag
\\
&\quad={\sf N}(t_{n+1};t_n+(\tau_{n+1}-\tau_n)v_n,\sigma_t^2)
\end{align}
where ${\sf N}(\,\cdot\,;\mu,\Sigma)$ denotes a normal distribution with mean $\mu$ and variance $\Sigma$, and $\sigma_v$ and $\sigma_t$ are standard deviations representing the degree of tempo variations and onset time fluctuations, respectively.
An initial distribution for $v_1$ is described similarly as a Gaussian ${\sf N}(v_1;v_{\rm ini},\sigma_{v,{\rm ini}}^2)$.

An algorithm to estimate onset score times and local tempos can be obtained by maximising the posterior probability $P(\bm\tau,\bm v|\bm t)\propto P(\bm t,\bm v|\bm\tau)P(\bm\tau)$.
This can be done by a standard Viterbi algorithm after discretisation of the tempo variables \cite{NakamuraEUSIPCO2016,Krebs2015}.
Note that this method does not use the pitch and offset information to estimate onset score times, which is typical in conventional onset rhythm transcription methods.
Since the period $G$ and rhythmic properties encoded in $P(s_n|s_{n-1})$ are dependent on the metre, in practice it is effective to consider multiple metrical HMMs corresponding to different metres, such as duple metre and triple metre, and choose the one with the maximum posterior probability in the stage of inference.

\subsection{Markov Random Field Model}\label{sec:Model}

Here we describe our main model.
As explained in Sec.~\ref{sec:Observation}, it is essential to combine a score model that enables prediction of note values given the input information of onset score times and pitches and a performance model that relates note values to actual durations realised in music performances.
To enable tractable inference and efficient parameter estimation, one should typically decompose each model into component models that involve a smaller number of stochastic variables.

As a framework to combine such component models, we consider the following Markov random field (MRF):
\begin{align}
&P(\bm r|\bm p,\bm d,\bar{\bm d},\bm\tau,\bm v)
\notag
\\
&\propto{\rm exp}\bigg[-\sum^N_{n=1}H_1(r_n;\bm\tau,\bm p)-\sum_{(n,m)\in\mathscr{N}}H_2(r_n,r_m)
\notag
\\
&\qquad\quad~~-\sum^N_{n=1}H_3(r_n;d_n,\bar{d}_n,v_n)\bigg].
\label{eq:MRF}
\end{align}
Here $H_1$ (called the {\it context model}) represents the prior model for each note value that depends on the onset score times and pitches, $H_2$ (the {\it interdependence model}) represents the interdependence of neighbouring pairs of note values ($\mathscr{N}$ denotes the set of neighbouring note pairs specified later) and $H_3$ (the {\it performance model}) represents the likelihood model.
Each term can be interpreted as an energy function that has small values when the arguments have higher probabilities.
The explicit forms of these functions are given as follows:
\begin{align}
H_1&=-\beta_1{\rm ln}\,P(r_n;\bm\tau,\bm p),
\\
H_2&=-\beta_2{\rm ln}\,P(r_n,r_m),
\\
H_3&=-\beta_{31}{\rm ln}\,P(d_n;r_n,v_n)-\beta_{32}{\rm ln}\,P(\bar{d}_n;r_n,v_n).
\label{eq:H3}
\end{align}
Each energy function is constructed with a negative log probability function multiplied by a positive weight.
These weights $\beta_1$, $\beta_2$, $\beta_{31}$ and $\beta_{32}$ are introduced to represent the relative importance of the component models.
For example, if we take $\beta_1=\beta_{31}=\beta_{32}=1$ and $\beta_2=0$, the model reduces to a Naive Bayes model with the durations considered as features.
For other values of $\beta$\,s, the model is no longer a generative model for the durations but still a generative model for the note values, which are the only unknown variables in our problem.
In the following we explain the component models in detail.
Learning parameters including $\beta$\,s is discussed in Sec.~\ref{sec:Learning}.

\subsubsection{Context Model}

The context model $H_1$ describes a prior distribution for note values that is conditionally dependent on given onset score times and pitches.
To construct this model, one should first specify the sample space of $r_n$, or, the set of possible values that each $r_n$ can take.
Based on the observations in Sec.~\ref{sec:Observation}, we consider the first ten IONVs as possible values of $r_n$.
Since $r_n$ can take other values in reality, we also consider a formally defined value `{\sf other}', which represents all other values of $r_n$.
Let
\[
\Omega_r(n)=\{{\rm IONV}(n,1),\ldots,{\rm IONV}(n,10),{\sf other}\}
\]
denote the sample space.
Therefore $P(r_n;\bm\tau,\bm p)$ is considered as an 11-dimensional discrete distribution.

\begin{figure}[t]
\begin{center}
\includegraphics[clip,width=0.95\columnwidth]{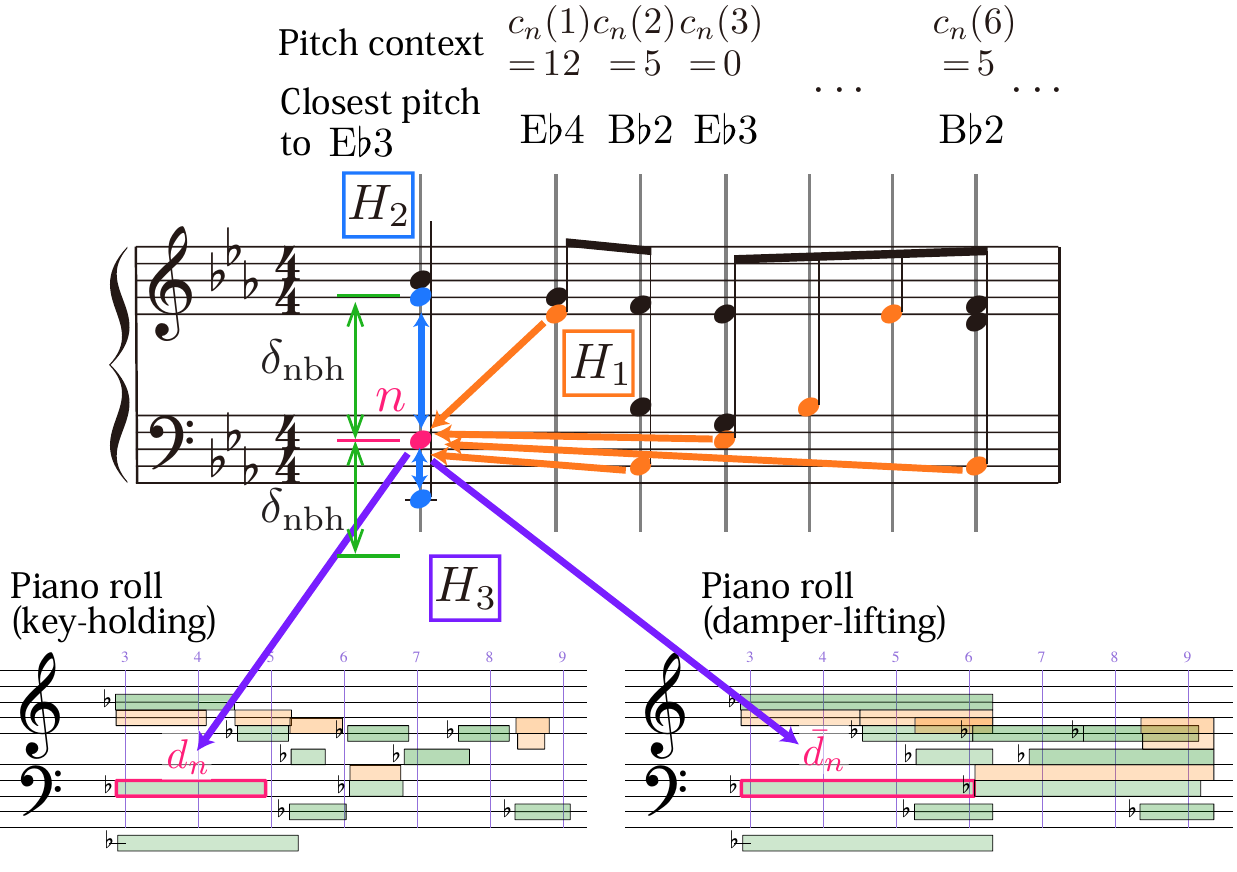}
\end{center}
\vspace{-5mm}
\caption{Statistical dependencies in the Markov random field model.}
\label{fig:MRF_Detail}
\vspace{-3mm}
\end{figure}
As we saw in Sec.~\ref{sec:Observation}, the distribution $P(r_n;\bm\tau,\bm p)$ depends heavily on the pitch context.
Based on our intuition that for each note $n$ the succeeding notes with a close pitch are most influential on the voice structure, in this paper we use the feature vector $c_n=(c_n(1),\ldots,c_n(10))$ as a context of note $n$, where $c_n(k)$ denotes the unsigned pitch interval between note $n$ and the closest pitch in its $k$-th next onset cluster.
An example of the context is given in Fig.~\ref{fig:MRF_Detail}.
Thus we have
\begin{equation}
P(r_n;\bm\tau,\bm p)=P(r_n;c_n(1),\ldots,c_n(10)).
\label{eq:ContextProbability}
\end{equation}
We remark that in general we can additionally consider different features (for example, metrical features) and our formulation in this section and in Sec.~\ref{sec:Learning} is valid independently of our particular choice of features.

Due to the huge number of different contexts for notes, it is not practical to use Eq.~(\ref{eq:ContextProbability}) directly.
With 88 pitches on a piano keyboard, each $c_n(k)$ can take 87 values and thus the right-hand side (RHS) of Eq.~(\ref{eq:ContextProbability}) has $11\cdot87^{10}$ parameters (or slightly less free parameters after normalisation), which is computationally infeasible.
(If one uses additional features, the number of parameters increases further.)
To solve this problem, we use a context-tree model \cite{ContextTreeClustering,Takaki2014}, in which contexts are categorised according to a set of criteria that are represented as a tree (as in decision tree analysis) and all contexts in one category have the same probability distribution.

\begin{figure}[t]
\begin{center}
\includegraphics[clip,width=0.6\columnwidth]{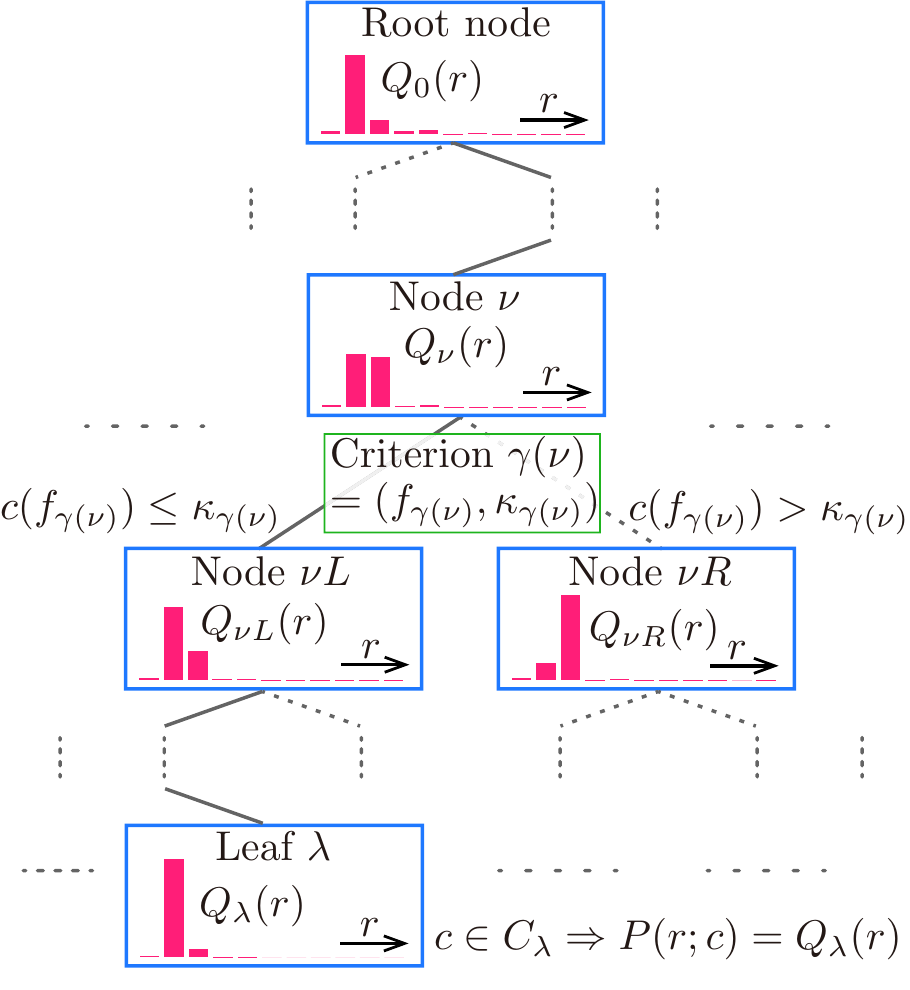}
\end{center}
\vspace{-5mm}
\caption{In a context-tree model, the distribution of a quantity $r$ is categorised with a set of criteria on the context $c$.}
\label{fig:ContextTreeModel}
\vspace{-3mm}
\end{figure}
Formally, a context-tree model is defined as follows.
Here we consider a general {\it context} $c=(c(1),\ldots,c(F))$, which is an $F$-dimensional feature vector.
We assume that the set of possible values for $c(f)$ is an ordered set for all $f=1,\ldots,F$ and denote it by $R_f$.
Let us denote the leaf nodes of a binary tree $T$ by $\partial T$.
Each node $\nu\in T$ is associated with a set of contexts denoted by $C_\nu$.
In particular, for the root node $0\in T$, $C_0$ is the set of all contexts ($R_1\times\cdots\times R_F$).
Each internal node $\nu\in T\setminus\partial T$ is associated with a criterion $\gamma(\nu)$ for selecting a subset of $C_\nu$.
A {\it criterion} $\gamma=(f_\gamma,\kappa_\gamma)$ is defined as a pair of a feature dimension $f_\gamma\in\{1,\ldots,F\}$ and a cut $\kappa_\gamma\in R_{f_\gamma}$.
The criterion divides a set of contexts $C$ into two subsets as
\begin{align}
C^L(\gamma)&=\{c\in C\,|\,c(f_\gamma)\leq\kappa_\gamma\},
\\
C^R(\gamma)&=\{c\in C\,|\,c(f_\gamma)>\kappa_\gamma\},
\end{align}
so that $C^L(\gamma)\cap C^R(\gamma)=\phi$ and $C^L(\gamma)\cup C^R(\gamma)=C$.
Now denoting the left and right child of $\nu\in T\setminus\partial T$ by $\nu L$ and $\nu R$, their sets of contexts are defined as $C_{\nu L}=C_\nu\cap C_0^L(\gamma(\nu))$ and $C_{\nu R}=C_\nu\cap C_0^R(\gamma(\nu))$, which recursively defines a {\it context tree} $(T,f,\kappa)$ (Fig.~\ref{fig:ContextTreeModel}).
By definition, a context is associated to a unique leaf node: for all $c\in C_0$ there exists a unique $\lambda\in\partial T$ such that $c\in C_\lambda$.
We denote such a leaf by $\lambda(c)$.
Finally, for each node $\nu\in T$, a probability distribution $Q_\nu(\,\cdot\,)$ is associated.
Now we can define the probability $P_\mathscr{T}(\,\cdot\,;c)$ as
\begin{equation}
P_\mathscr{T}(\,\cdot\,;c)=Q_{\lambda(c)}(\,\cdot\,).
\end{equation}
The tuple $\mathscr{T}=(T,f,\kappa,Q)$ defines a {\it context-tree model}.

For a context-tree model with $L$ leaves, the number of parameters for the distribution of note values is now reduced to $11L$.
In general a model with a larger tree size has more ability to approximate Eq.~(\ref{eq:ContextProbability}) at the cost of an increasing number of model parameters.
The next problem is to find the optimal tree size and the optimal criterion for each internal node.
We will explain this in Sec.~\ref{sec:ContextTreeClustering}.

\subsubsection{Interdependence Model}

Although the distribution of note values in the context model is dependent on the pitch context, it is independently defined for each note value.
As explained in Sec.~\ref{sec:Observation}, interdependence of note values is also important since it arises from logical constraint on the voice structure.
Such interdependence can be described with a joint probability of note values of a pair of notes in $H_2$.
As in the context model, we consider the set $\Omega_r$ as a sample space for note values so that the joint probability $P(r_n,r_m)$ for notes $n$ and $m$ has $11^2$ parameters.

The choice of the set of neighbouring note pairs $\mathscr{N}$ in Eq.~(\ref{eq:MRF}) is most important for the interdependence model.
In order to capture the voice structure we define $\mathscr{N}$ as
\begin{equation}
\mathscr{N}=\{(n,m)\,|\,\tau_n=\tau_m,~|p_n-p_m|\leq \delta_{\rm nbh}\}
\label{eq:Neighbour}
\end{equation}
where $\delta_{\rm nbh}$ is a parameter to define the vicinity of the pitch.
The value of $\delta_{\rm nbh}$ is determined from data (see Sec.~\ref{sec:Optimisation}).

\subsubsection{Performance Model}\label{sec:PerformanceModel}

The performance model is constructed with the probability of actual durations in performances given a note value and a local tempo.
Since we can use two durations $d_n$ and $\bar{d}_n$, two distributions, $P(d_n;r_n,v_n)$ and $P(\bar{d}_n;r_n,v_n)$, are considered for each note as in the RHS of Eq.~(\ref{eq:H3}).
To regulate the effect of varying tempos and avoid the increase in the complexity of the model to handle possibly many types of note values, we consider distributions over normalised durations, $d'_n=d_n/(r_nv_n)$ and $\bar{d}'_n=\bar{d}_n/(r_nv_n)$, as we did in Sec.~\ref{sec:Observation}.
We therefore assume
\begin{equation}
P(d_n;r_n,v_n)=g(d'_n)\quad{\rm and}\quad P(\bar{d}_n;r_n,v_n)=\bar{g}(\bar{d}'_n)
\label{eq:PerfmModel}
\end{equation}
where $g$ and $\bar{g}$ are one-dimensional probability distributions supported on positive real numbers.

The histograms corresponding to $g$ and $\bar{g}$ taken from performance data described in Sec.~\ref{sec:Learning} are illustrated in Fig.~\ref{fig:Distribution}.
One can recognise two (one) peak(s) for the distribution of normalised key-holding (damper-lifting) durations.
Since theoretical forms of these distributions are unknown, we use as phenomenologically fitting distributions the following generalised inverse-Gaussian (GIG) distribution:
\begin{equation}
{\sf GIG}(x;a,b,h)=\frac{(a/b)^{h/2}}{2K_h(2\sqrt{ab})}x^{h-1}e^{-(ax+b/x)}
\end{equation}
where $a,b>0$ and $h\in\mathbb{R}$ are parameters and $K_h$ denotes the modified Bessel function of the second kind.
The GIG distributions are supported on positive real numbers and include the gamma ($a\to0$), inverse-gamma $(b\to0$) and inverse-Gaussian ($h=-1/2$) distributions as special cases.
Since a GIG distribution has only one peak, we use a mixture of GIG distributions to represent $g$.
We parameterise $g$ and $\bar{g}$ as
\begin{align}
g(x)&=w_1{\sf GIG}(x;a_1,b_1,h_1)+w_2{\sf GIG}(x;a_2,b_2,h_2),
\label{eq:g}
\\
\bar{g}(x)&={\sf GIG}(x;a_3,b_3,h_3)
\label{eq:gbar}
\end{align}
where $w_1$ and $w_2=1-w_1$ are mixture weights.
Parameter values obtained from data are given in Sec.~\ref{sec:PerformanceModelLearning}.

\subsection{Model Learning}\label{sec:Learning}

Similarly as the language model and the acoustic model for a speech recognition system are generally trained separately with different data, our three component models can be trained separately and combined afterwards to determine the optimal weights (the $\beta$\,s).
The context model and the interdependence model can be learned with musical score data and we used a dataset of 148 classical piano pieces (with $3.4\times10^6$ notes) by various composers\footnote{The lists of used pieces for the score data and the performance data are available at the accompanying web page \cite{Webpage}.}.
On the other hand, the performance model requires performance data aligned with reference scores.
The used data consisted of 180 performances (60 phrases $\times$ 3 different players) by various composers and various performers that are mostly collected from publicly available MIDI performances recorded in international piano competitions \cite{eCompetition}.
Due to the lack of abundant data, we used the same performance data for training and evaluation.
Because the number of parameters for the performance model is small (ten independent parameters in $g$ and $\bar{g}$ and two weight parameters) and they are not fine-tunable, there should be little concern about overfitting here and most comparative evaluations in Sec.~\ref{sec:Evaluation} are done with equal conditions.
(See also the discussion in Secs.~\ref{sec:PerformanceModelLearning} and \ref{sec:Examinations}.)
To avoid overfitting, the score data and the performance data contained no overlapping musical pieces (at least in units of movements).
Learning methods for the component models are described in the following sections and Sec.~\ref{sec:Optimisation} describes the optimisation of the $\beta$\,s.

\subsubsection{Learning the Context Model}\label{sec:ContextTreeClustering}

The context-tree model can be learned by growing the tree based on the maximum likelihood (ML) principle, which is called {\it context-tree clustering}.
This is usually done by recursively splitting a node that minimises the likelihood \cite{ContextTreeClustering}.
Although it is not essentially new, we describe the learning method here for the readers' convenience because context-tree clustering is not commonly used in the field of music informatics and in articles for speech processing (where it is widely used) the notations are adapted for the case with Gaussian distributions, which is not ours.

Let $x_i=(r_i,c_i)$ denote a sample extracted from score data, where $i$ denotes a note in the score data, $r_i$ denotes an element in $\Omega_r(i)$ and $c_i$ denotes the context of note $i$.
The set of all samples will be denoted by $\bm x=(x_i)_{i=1}^I$.
The log likelihood $L_\mathscr{T}(\bm x)$ of a context-tree model $\mathscr{T}=(T,f,\kappa,Q)$ is given as
\begin{align}
L_\mathscr{T}(\bm x)&=\sum_{i=1}^I{\rm ln}\,P_\mathscr{T}(x_i)=\sum_{i=1}^I{\rm ln}\,Q_{\lambda(c_i)}(x_i)
\notag
\\
&=\sum_{\lambda\in\partial T}\sum_{i:\,c_i\in C_\lambda}q_\lambda(x_i)
\end{align}
where in the second line we decomposed the samples according to the criteria of the leaves and hereafter we denote $q_\nu(\,\cdot\,)={\rm ln}\,Q_\nu(\,\cdot\,)$ for each node $\nu$.
The parameters for each distribution $Q_\nu$ for node $\nu\in T$ are learned from the samples $\{x_i|c_i\in C_\nu\}$ according to the ML method.
We implicitly understand that all $Q$\,s are already learned in this way.

Given a context tree $\mathscr{T}^{(m)}$ (one begins with a tree $\mathscr{T}^{(0)}$ containing only the root node and proceeds $m=0,1,2,\ldots$ as follows), one of the leaves $\lambda\in\partial T^{(m)}$ is split according to some additional criterion $\gamma(\lambda)$.
Let us denote the expanded context-tree model by $\mathscr{T}^{(m)}_\lambda$.
Since $\mathscr{T}^{(m)}_\lambda$ is same as $\mathscr{T}^{(m)}$ except for the new leaves $\lambda L$ and $\lambda R$, the difference of log likelihoods $\Delta L(\lambda)=L_{\mathscr{T}^{(m)}_\lambda}(\bm x)-L_{\mathscr{T}^{(m)}}(\bm x)$ is given as
\begin{equation}
\sum_{i:\,c_i\in C_{\lambda L}}q_{\lambda L}(x_i)+\sum_{i:\,c_i\in C_{\lambda R}}q_{\lambda R}(x_i)-\sum_{i:\,c_i\in C_\lambda}q_{\lambda}(x_i).
\end{equation}
Note that $\Delta L(\lambda)\geq0$ since $Q_{\lambda L}$ and $Q_{\lambda R}$ have the ML.
Now the leaf $\lambda^*$ and the criterion $\gamma(\lambda^*)$ that maximise $\Delta L(\lambda)$ are selected for growing the context tree: $\mathscr{T}^{(m+1)}=\mathscr{T}^{(m)}_{\lambda^*}$.

According to the above ML criterion, the context tree can be expanded to the point where all samples are completely separated by contexts, for which the model often suffers from overfitting.
To avoid this and find an optimal tree size according to the data, the minimal description length (MDL) criterion for model selection can be used \cite{Shinoda2000,Rissanen1984}.
The MDL $\ell_{\cal M}(\bm x)$ for a model ${\cal M}$ with parameters $\theta_{\cal M}$ is given as
\begin{equation}
\ell_{\cal M}(\bm x)=-{\rm log}_2P(\bm x;\hat{\theta}_{\cal M})+\frac{|{\cal M}|}{2}{\rm log}_2I
\label{eq:MDL}
\end{equation}
where $I$ is the length of $\bm x$, $|{\cal M}|$ is the number of free parameters of model ${\cal M}$ and $\hat{\theta}_{\cal M}$ denotes the ML estimate of $\theta_{\cal M}$ according to data $\bm x$.
Here, the first term in the RHS is the negative log likelihood, which in general decreases when the model's complexity increases.
On the other hand, the second term increases when the number of model parameters increases.
Thus a model that minimises the MDL is chosen by a trade off of the model's precision and complexity.
The MDL criterion is justified by an information-theoretic argument \cite{Rissanen1984}.

For our context-tree model, each $Q$ is an 11-dimensional discrete distribution and has ten free parameters, and therefore the increase of parameters by expanding a node is ten.
Substituting this into Eq.~(\ref{eq:MDL}), we find
\begin{align}
\Delta\ell(\lambda^*)&=\ell_{\mathscr{T}^{(m+1)}}(\bm x)-\ell_{\mathscr{T}^{(m)}}(\bm x)
\notag
\\
&=-\Delta L(\lambda^*)/({\rm ln}\,2)+(10/2){\rm log}_2I.
\end{align}
In summary, the context tree is expanded by splitting the optimal leaf $\lambda^*$, up to a step where $\Delta\ell(\lambda^*)$ becomes positive.

\begin{figure}[t]
\begin{center}
\includegraphics[clip,width=1.\columnwidth]{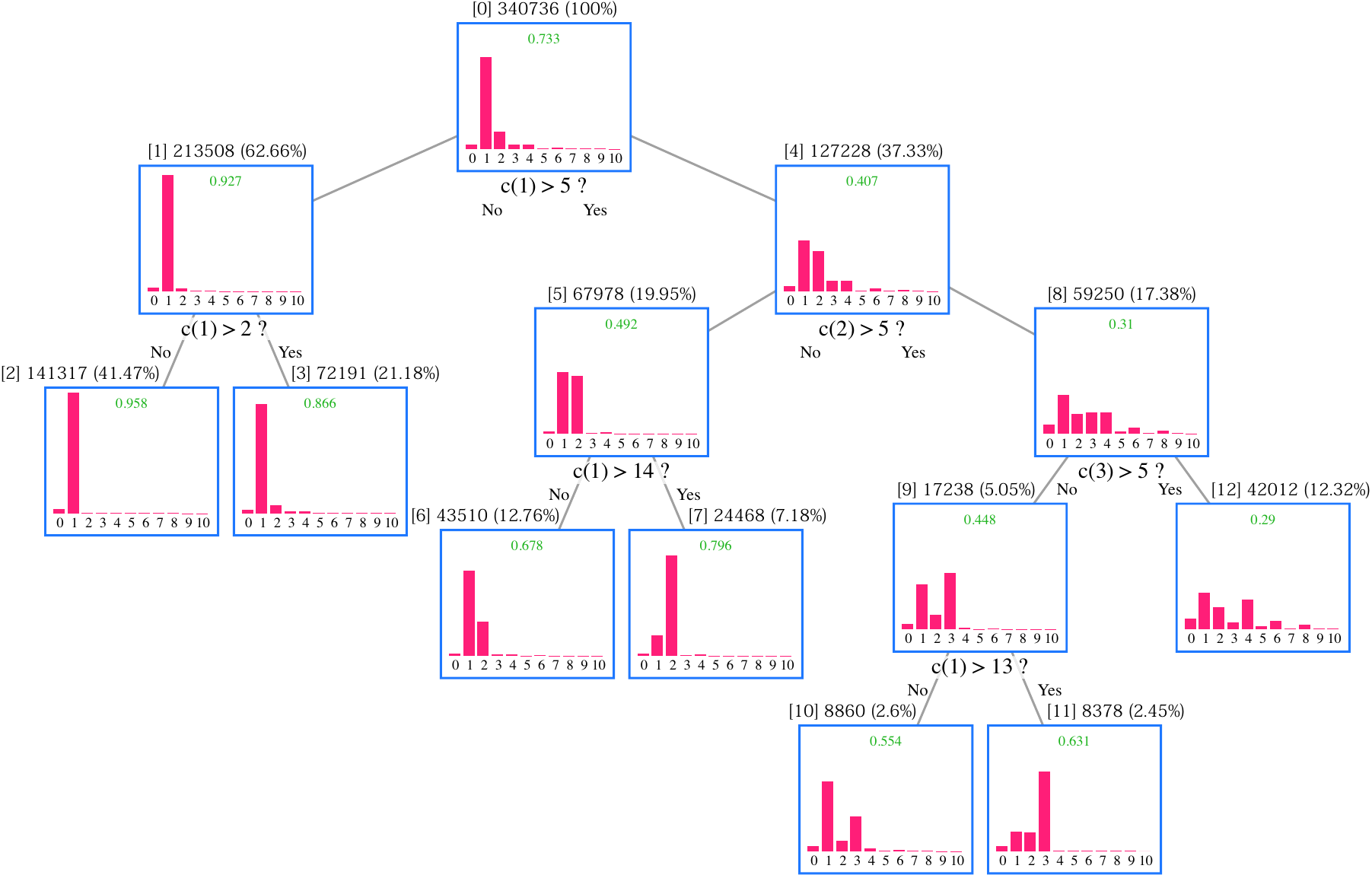}
\end{center}
\vspace{-3mm}
\caption{A subtree of the obtained context-tree model. Above each node are indicated the node ID, number of samples and their proportion in the whole data and the green number indicates the highest probability in each distribution. See text for explanation of the labels for each distribution.}
\label{fig:ContextTree}
\vspace{-3mm}
\end{figure}
With our score data of $3.4\times10^6$ musical notes, the learned context tree had 132 leaves.
A subtree is illustrated in Fig.~\ref{fig:ContextTree} where the node ID is shown in square brackets and the labels $1,\ldots,10$ in the distribution show those probabilities correspond to ${\rm IONV}(1),\ldots,{\rm IONV}(10)$ and the label 0 is assigned to the `{\sf other}'.
For example, one finds a distribution with a sharp peak at ${\rm IONV}(1)$ in node 2 whose contexts satisfy $c(1)\leq 2$.
This can be interpreted as follows: if note $n$ has a pitch within 2 semitones in the next onset cluster, then it is highly probable that they are in the same voice and note $n$ has $r_n={\rm IONV}(n,1)$.
On the other hand, the ${\rm IONV}(2)$ has the largest probability in node 7 (the distribution is the same one as in Fig.~\ref{fig:DistrIONVClass}(d)) with contexts satisfying $c(2)\leq 5$ and $c(1)>14$, whose interpretation was explained in Sec.~\ref{sec:HintsForScoreModel}.　
Similar interpretations can be made for node 11 and other nodes.
These results show that the context tree tries to capture the voice structure through the pitch context.
As this is induced from data in an unsupervised way, it serves as an information-scientific confirmation that the voice structure has a strong influence on the configuration of note values.

\subsubsection{Learning the Interdependence Model}\label{sec:LearningH2}

\begin{figure}[t]
\begin{center}
\includegraphics[clip,width=0.85\columnwidth]{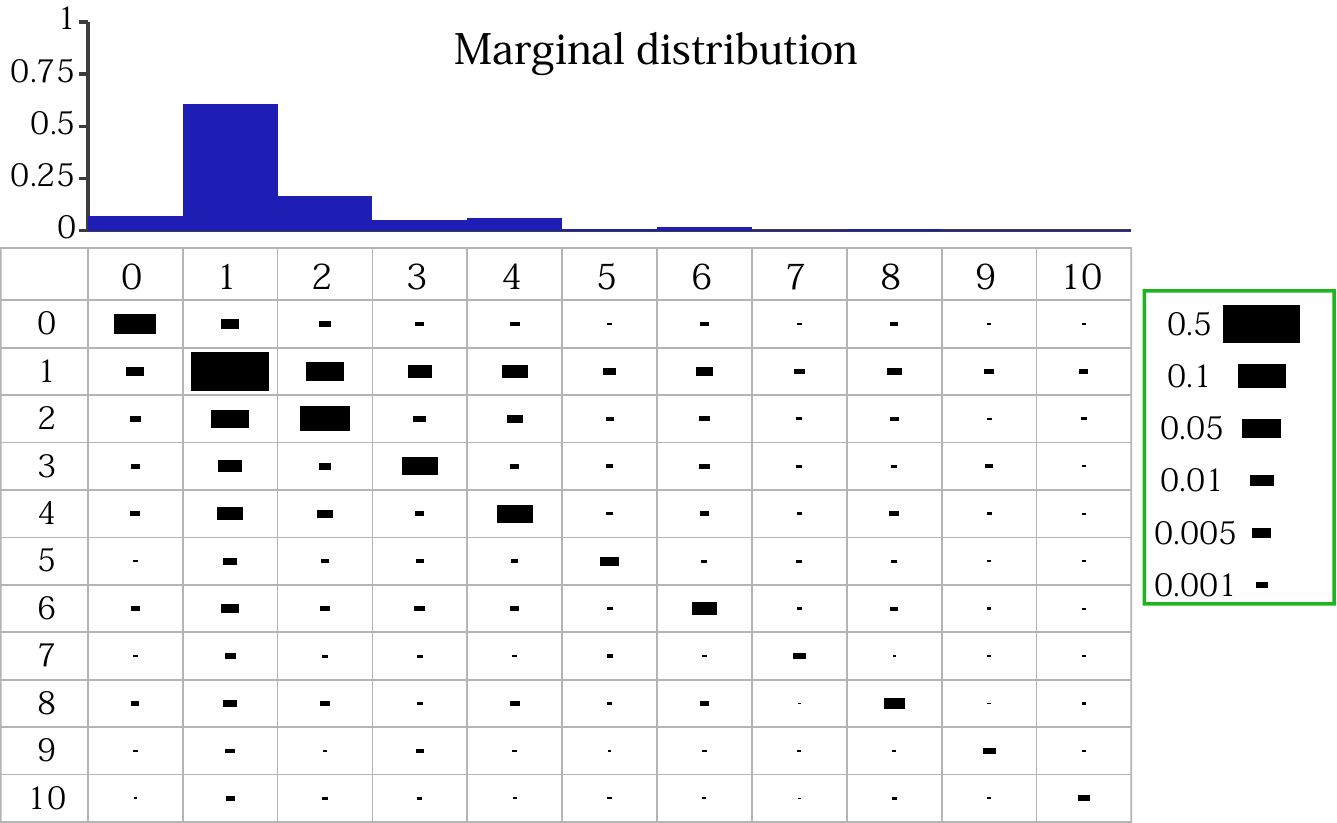}
\end{center}
\vspace{-3mm}
\caption{Joint probability distribution of note values obtained for the interdependence model for $\delta_{\rm nbh}=12$. See text for explanation of the labels.}
\label{fig:InterDepIONVProb}
\vspace{-3mm}
\end{figure}

The interdependence model for each $\delta_{\rm nbh}$ can be directly learned from score data: for all note pairs defined by Eq.~(\ref{eq:Neighbour}), one obtains the joint probability of their note values.
The obtained results for $\delta_{\rm nbh}=12$ is shown in Fig.~\ref{fig:InterDepIONVProb} where the same labels are used as in Fig.~\ref{fig:ContextTree}.
The diagonal elements, which have the largest probability in each row and column, clearly reflect the constraint of chordal notes having the same note values.

Since the interdependence model is by itself not as precise a generative model as the context model and these models are not independent, we optimise $\delta_{\rm nbh}$ in combination with the context model.
This is described in Sec.~\ref{sec:Optimisation}, together with the optimisation of the weights.
In preparation for this, we learned the joint probability for each of $\delta_{\rm nbh}=0,1,\ldots,15$.

\subsubsection{Learning the Performance Model}\label{sec:PerformanceModelLearning}

\begin{figure}[t]
\begin{center}
\subfigure[]
{\includegraphics[clip,width=0.85\columnwidth]{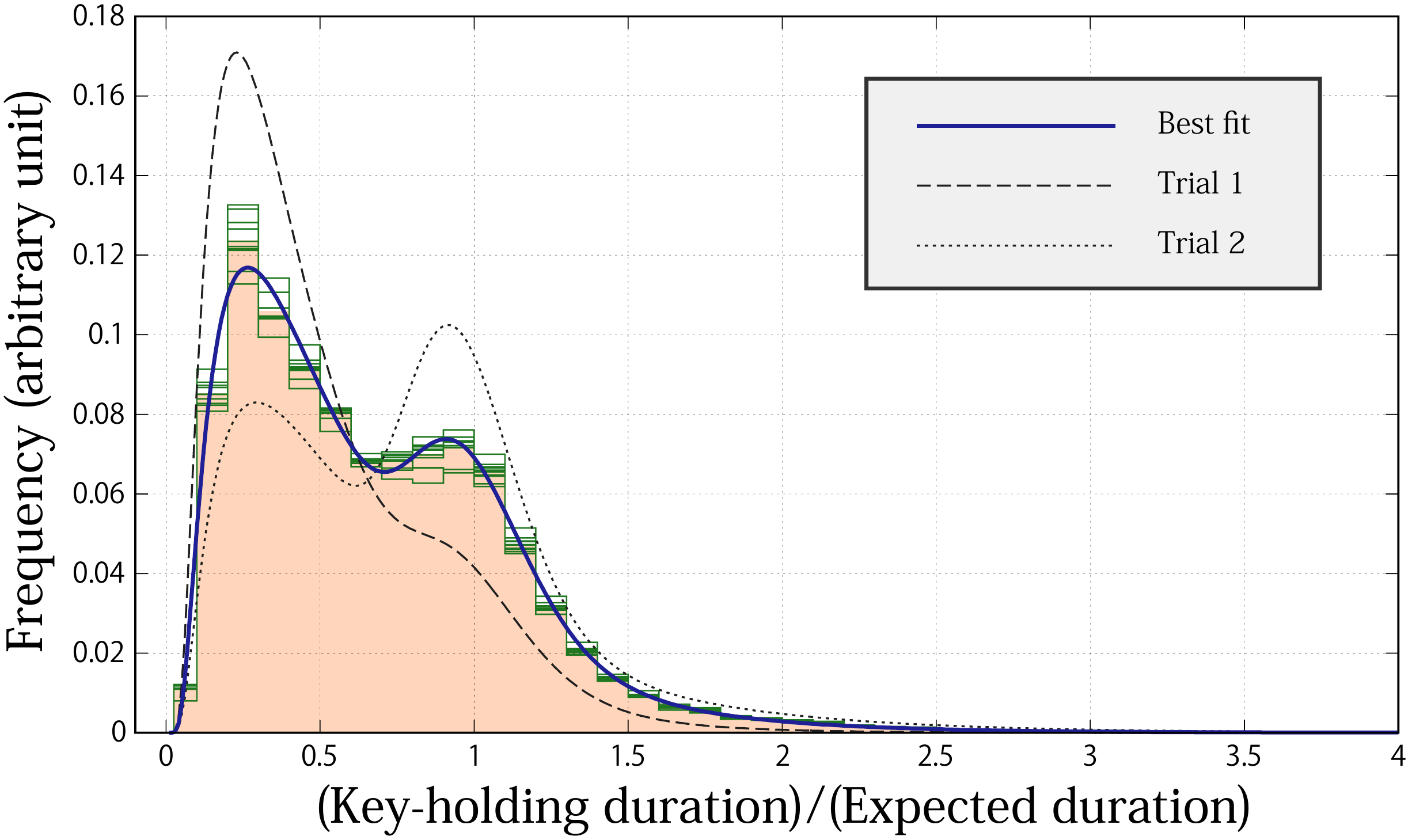}}\\
\subfigure[]
{\includegraphics[clip,width=0.85\columnwidth]{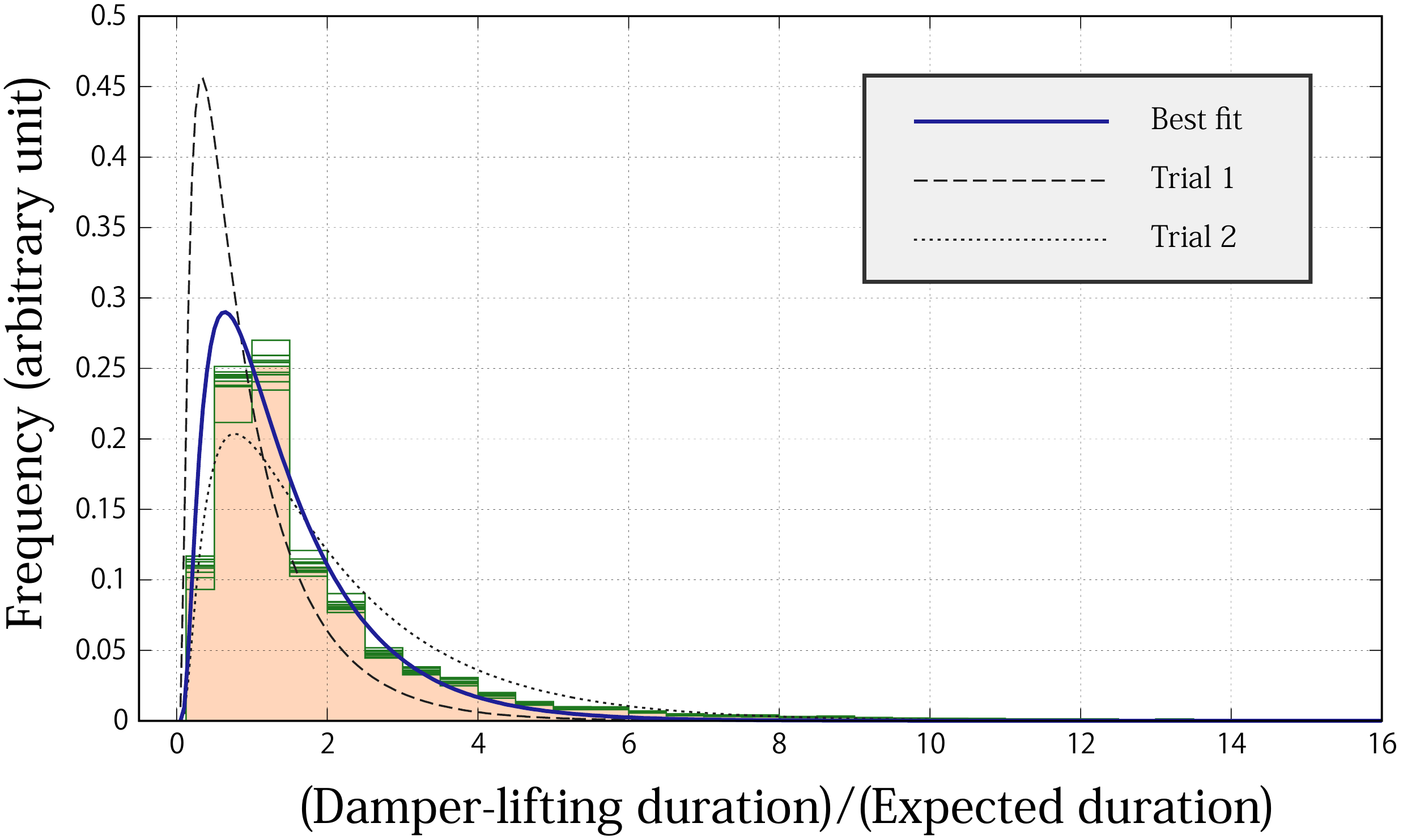}}
\vspace{-3mm}
\end{center}
\caption{Distributions used for the performance model for (a) key-holding durations and (b) damper-lifting durations. In each figure, the background histogram is the one obtained from the whole training data (same as Fig.~\ref{fig:Distribution}) and the superposed histograms are obtained from 10-fold training datasets.}
\label{fig:VarPerfmModel}
\vspace{-3mm}
\end{figure}
The parameters for the performance model in Eqs.~(\ref{eq:g}) and (\ref{eq:gbar}) are learned from the distributions given in Fig.~\ref{fig:Distribution}.
We performed a grid search for minimising the squared fitting error for each distribution.
The obtained values are the following:
\begin{align}
a_1&=2.24\pm0.02,\quad b_1=0.24\pm0.01,\quad h_1=0.69\pm0.01,
\notag
\\
a_2&=13.8\pm0.1,\quad b_2=15.2\pm0.1,\quad h_2=-1.22\pm0.04,
\notag
\\
w_1&=0.814\pm0.004,\quad w_2=0.186\pm0.004,
\notag
\\
a_3&=0.94\pm0.01,\quad b_3=0.51\pm0.01,\quad h_3=0.80\pm0.01.
\notag
\end{align}
The fitting curves are illustrated in Fig.~\ref{fig:VarPerfmModel}.
In the figure, we also show histograms of normalised durations obtained from ten different subsets of the training data that are constructed similarly as the 10-fold cross-validation method: i.e.\ we split the training data into ten separate sets (each containing 10\% of the performances) and the remaining 90\% of the data were used as one of the 10-fold training datasets.
We can see in Fig.~\ref{fig:VarPerfmModel} that the differences among these histograms are not large.
Two other parameter sets for $g$ and $\bar{g}$ were chosen as trial distributions shown in the figure, which deviate from the best fit distribution more than the differences among the 10-fold histograms.
These distributions are used in Sec.~\ref{sec:InfluenceOfPerfmModel} to examine the influence of the parameter values for the performance model.

\subsubsection{Optimisation of the Weights}\label{sec:Optimisation}

Since the three component models for the MRF model in Eq.~(\ref{eq:MRF}) are not independent, the weights $\beta$ should be obtained by simultaneous optimisation using performance data in general.
However, since the amount of score data at hand is significantly larger than that of the performance data, we optimise the weights in a more efficient way.
Namely, we first optimise $\beta_1$ and $\beta_2$ with the score data and then optimise $\beta_{31}$ and $\beta_{32}$ with the performance data (with fixed $\beta_1$ and $\beta_2$).
When examining the influence of varying these weights in Sec.~\ref{sec:Examinations}, we will discuss that the influence of this sub-optimisation procedure is seemingly small.

We obtained the first two weights simultaneously with $\delta_{\rm nbh}$ by the ML principle with the following results:
\begin{equation}
\hat{\beta}_1=0.965\pm0.005,\quad\hat{\beta}_2=0.03\pm0.005,\quad\hat{\delta}_{\rm nbh}=12.
\label{eq:UsedBeta12}
\end{equation}
The result $\hat{\beta}_2\ll\hat{\beta}_1$ indicates that the interdependence model has little influence in the score model.
Although it seems somewhat contradictory to the results in Sec.~\ref{sec:LearningH2} at first sight, we can understand this by noticing that both the context model and interdependence model make use of pitch proximity to capture the voice structure.
The former model uses pitch proximity in the horizontal (time) direction and the latter model does so in the vertical (pitch) direction, and they have overlapping effects since whenever a note pair (say, note $n$ and $n'$) in an onset cluster have close pitches, they tend to share notes in succeeding onset clusters with close pitches (see e.g.\ the chords in the left-hand part in the score in Fig.~\ref{fig:ExResult}).
Thus note $n$ and $n'$ tend to obey similar distributions in the context model.
Since the interdependence model is weaker in terms of predictive ability, this results in small $\hat{\beta}_2$.

We optimised $\beta_{31}$ and $\beta_{32}$ according to the accuracy of note value recognition (more precisely, the average error rate defined in Sec.~\ref{sec:Measure}) and the obtained values are as follows:
\begin{equation}
\hat{\beta}_{31}=0.21\pm0.01,\quad\hat{\beta}_{32}=0.003\pm0.001.
\label{eq:UsedBeta3132}
\end{equation}
One can notice that $\hat{\beta}_{32}\ll\hat{\beta}_{31}$, which can be explained by the significantly larger variance of the distribution of damper-lifting durations than that of key-holding durations in Fig.~\ref{fig:Distribution}.
On the other hand, the result that $\hat{\beta}_{31}$ is considerably smaller than $\hat{\beta}_1$ can be interpreted as that the score model has more importance for estimating note values (in our model).
The effect of varying weights is examined in Sec.~\ref{sec:Examinations}.

\subsection{Inference Algorithm and Implementation}\label{sec:Implementation}

We can develop a note value recognition algorithm based on the maximisation of the probability in Eq.~(\ref{eq:MRF}) with respect to $\bm r$.
As a search space, we consider $\Omega_r(n)\setminus\{{\sf other}\}$ for each $r_n$.
Without $H_2$, the probability is independent for each $r_n$ and the optimisation is straightforward.
With $H_2$, we should optimise those $r_n$\,s connected in $\mathscr{N}$ simultaneously.
Since there are only vertical interdependencies in our model, the optimisation can be done independently for each onset cluster.
With $J$ notes in an onset cluster, the set of candidate note values has size $10^J$.
Typically $J\leq6$ for piano scores and the global search can be done directly.
Occasionally, however, $J$ can be ten or more and the computation time can be too large.
To reduce the size of search space in this case, cutoffs are placed on the order of IONVs when $J>6$ in our implementation: instead of the first ten IONVs, we use the first $(14-J)$ IONVs for $6 < J \leq 10$ and two IONVs for $J > 10$.
Although with this approximation we lose a certain proportion of possible solutions, we know that this proportion is small from the small probability of $r$ having higher IONVs in Fig.~\ref{fig:DistrIONVClass}(b).

Our implementation of the MRF model and the metrical HMM for onset rhythm transcription and tempo estimation is available \cite{Webpage}.
A tempo estimation algorithm based on a Kalman smoother is also provided for applying our method to results of other onset rhythm transcriptions that do not include tempo information as output.

\section{Evaluation}\label{sec:Evaluation}

\subsection{Evaluation Measures}\label{sec:Measure}

We first define evaluation measures used in our study.
For each note $n=1,\ldots,N$, let $r^{\rm c}_n$ and $r^{\rm e}_n$ be the correct and estimated note values.
Then the {\it error rate} ${\cal E}$ is defined as
\begin{equation}\label{eq:ErrorRate}
{\cal E}=\frac{1}{N}\sum_{n=1}^N\mathbb{I}(r^{\rm e}_n\neq r^{\rm c}_n)
\end{equation}
where $\mathbb{I}({\cal C})$ is 1 if condition ${\cal C}$ is true and 0 otherwise.
This measure does not take into account how close the estimation is to the correct value when they are not exactly equal.
Alternatively one can consider the averaged `distance' between the estimated and correct note values.
As such a measure we define the following {\it scale error} ${\cal S}$:
\begin{equation}\label{eq:NoteWiseMeasure}
{\cal S}={\rm exp}\bigg[\frac{1}{N}\sum_n|{\rm ln}(r^{\rm e}_n/r^{\rm c}_n)|\bigg].
\end{equation}
The difference and average is defined in the logarithmic domain to avoid bias for larger note values.
${\cal S}$ is unity if all note values are correctly estimated, and for example, ${\cal S}=2$ if all estimations are doubled or halved from the correct values.

Because of the ambiguity of defining the beat unit, score times estimated by rhythm transcription methods often have doubled, halved or other scaled values \cite{Nakamura2017,Cemgil2000B}, which should not be treated as complete errors.
To handle such scaling ambiguity, we normalise note values with the first IONV as
\begin{align}
r'^{\rm e}_n=r^{\rm e}_n/{\rm IONV}^{\rm e}(n,1),
\\
r'^{\rm c}_n=r^{\rm c}_n/{\rm IONV}^{\rm c}(n,1)
\end{align}
where ${\rm IONV}^{\rm e}(n,1)$ and ${\rm IONV}^{\rm c}(n,1)$ is the first IONV defined for the estimated and correct score, respectively.
Scale-invariant evaluation measures can be obtained by applying Eqs.~(\ref{eq:ErrorRate}) and (\ref{eq:NoteWiseMeasure}) for $r'^{\rm e}_n$ and $r'^{\rm c}_n$.

\subsection{Comparative Evaluations}\label{sec:Comparisons}

In this section, we evaluate the proposed method, a previously studied method \cite{Temperley2009} and a simple model discussed in Sec.~\ref{sec:Observation} on our data set and compare them in terms of the accuracy of note value recognition.

\subsubsection{Setup}

\begin{figure}[t]
\begin{center}
\subfigure
{\includegraphics[clip,width=0.95\columnwidth]{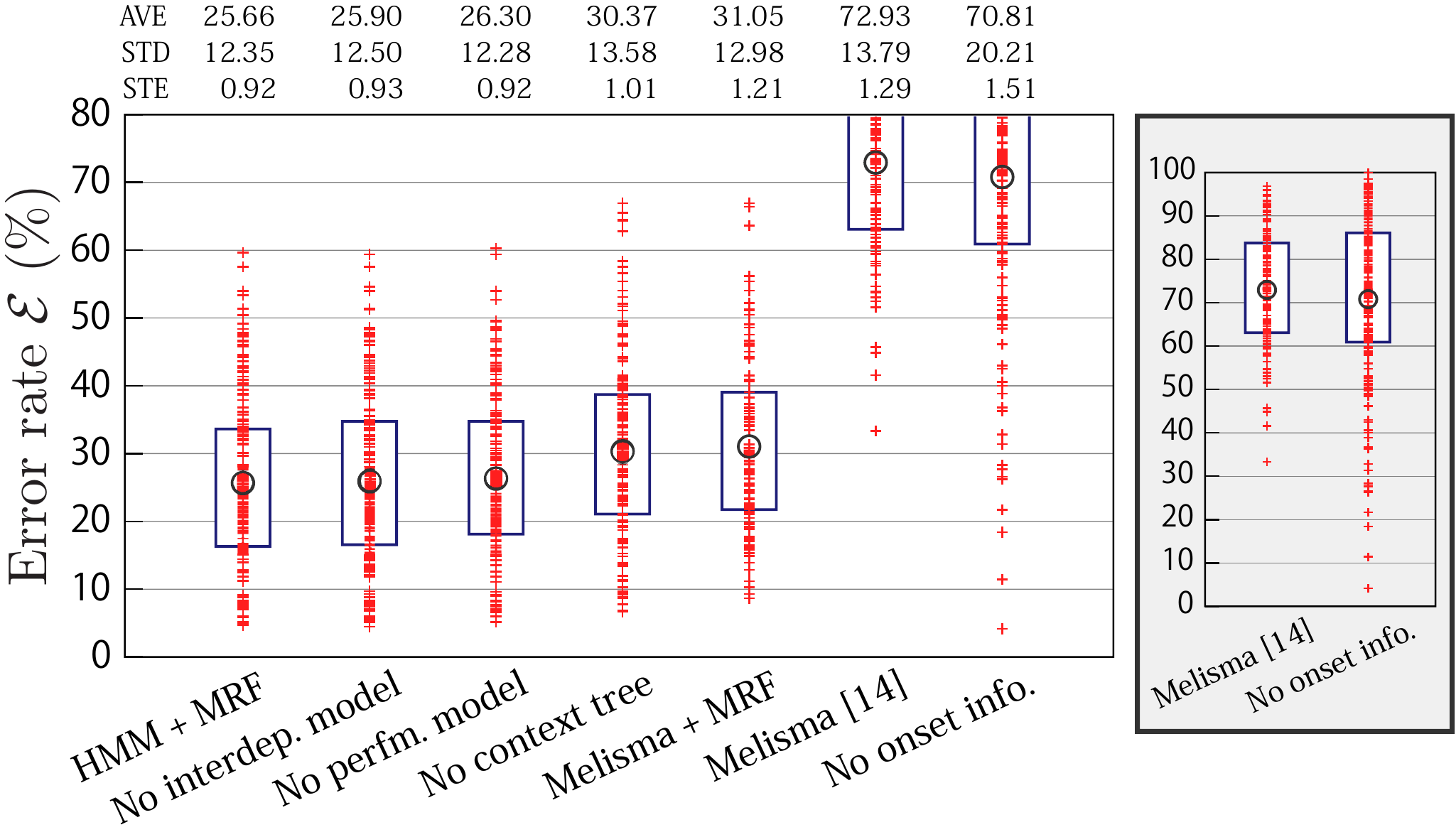}}
\\
\subfigure
{\includegraphics[clip,width=0.95\columnwidth]{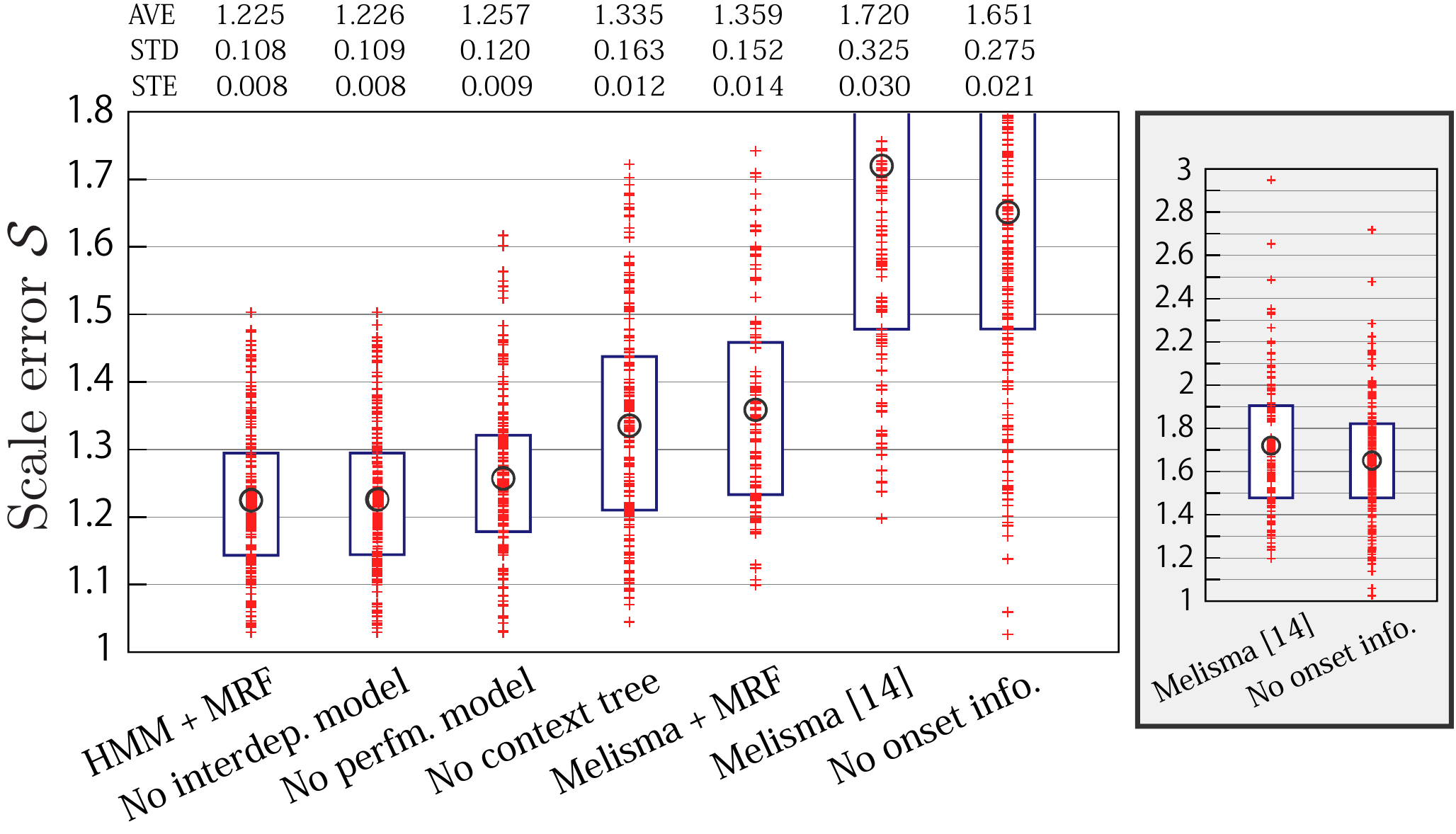}}
\end{center}
\vspace{-3mm}
\caption{Piece-wise average error rates and scale errors of note value recognition. Each red cross corresponds to one performance. The circle indicates the average (AVE) and the blue box indicates the range from the first to third quartiles, and STD and STE indicate the standard deviation and the standard error.}
\label{fig:ErrorRate}
\vspace{-3mm}
\end{figure}
To study the contribution of the component models of our MRF model, we evaluated the full model, a model without the interdependence model ($\beta_2=0$), a model without the performance model ($\beta_{31}=\beta_{32}=0$) and an MRF model with a context model having no (or a trivial) context tree, all applied to the result of onset rhythm transcription by the metrical HMM.
For the metrical HMM, we use the parameter values taken from a previous study \cite{Nakamura2017}.
These parameters were learned with the same score data and different performance data.

In addition, we evaluated a method based on a simple prior distribution on note values (Fig.~\ref{fig:DistrIONVClass}(a)) combined with an output probability $P(d_n;r_n,v_n)$ in Eq.~(\ref{eq:PerfmModel}), which uses no information of onset score times.
For comparison, we evaluated the Melisma Analyzer (version 2) \cite{Temperley2009}, which is to our knowledge the only major method that can estimate onset and offset score times, and we also applied post-processing by the proposed method on the onset score times obtained by the Melisma Analyzer.
The used data is described in Sec.~\ref{sec:Learning}.

\subsubsection{Results}

The piece-wise average error rates and scale errors are shown in Fig.~\ref{fig:ErrorRate} where the mean (AVE) over all pieces and the standard error for the mean (corresponding to $1\sigma$ deviation in the $t$-test) are also given.
Out of the 180 performances, only 115 performances were properly processed by the Melisma Analyzer and are depicted in the figure.
In addition, 30.0\% of the note values estimated by the method were zero and scale errors were calculated without these values.
One can see that the Melisma Analyzer and the simple model without using the onset score time information have high error rates and the proposed methods clearly outperformed them.

\begin{figure}[t]
\begin{center}
\includegraphics[clip,width=0.85\columnwidth]{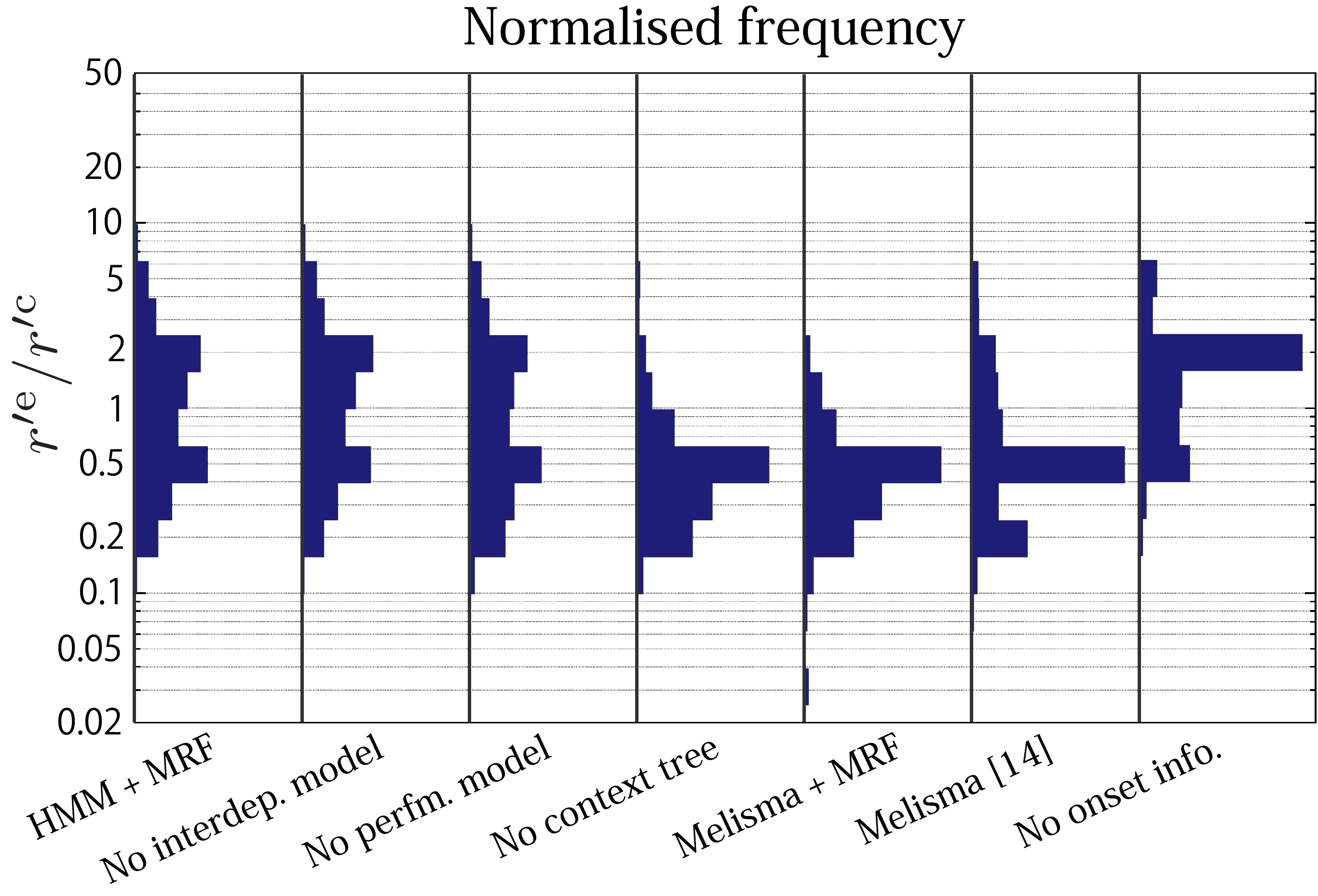}
\end{center}
\vspace{-3mm}
\caption{Distributions of note-wise scale errors $r'^{\rm e}/r'^{\rm c}$ for notes with $r'^{\rm e}/r'^{\rm c}\neq1$.}
\label{fig:NotewiseScaleError}
\end{figure}

The distributions of note-wise scale errors $r'^{\rm e}/r'^{\rm c}$ for incorrect estimations ($r'^{\rm e}/r'^{\rm c}\neq1$) in Fig.~\ref{fig:NotewiseScaleError} show that the Melisma Analyzer (simple model) more often estimates note values shorter (longer) than the correct ones.
For the simple model, this is because it mostly relies on, other than a relatively weak prior distribution in Fig.~\ref{fig:DistrIONVClass}(a), the distribution of key-holding durations in Fig.~\ref{fig:Distribution}(a), which has the highest peak position lower than its mean.
For the Melisma Analyzer, the short and zero note values arise because the method quantises the onset and (key-release) offset times into analysis frames of 50 ms.
Whereas the comparison is not fair in that the Melisma Analyzer can potentially identify grace notes with zero note values, which our data did not contain and our method cannot recognise, the rate (30.0\%) is considerably higher than their typical frequency in piano scores.

Among the different conditions for the proposed method, the full model had the best accuracy and the case with no context tree had significantly worse results, showing a clear effect of the context model.
Compared to the full model, the average error rate for the model without the performance model was worse but within $1\sigma$ deviation and the average scale error was significantly worse, indicating that the performance model has an effect in approximating the estimated note values to the correct ones.
On the other hand, results without the interdependence model were slightly worse but almost the same as the full model, which is because of the small $\hat{\beta}_2$.
The last result indicates that one can remove the interdependence model without much increase of estimation errors, which simplifies the inference algorithm as the distributions of note values become independent for each note.

\subsection{Examining the Proposed Model}\label{sec:Examinations}

Here we examine the proposed model in greater depth.

\subsubsection{Error Analyses}

\begin{figure}[t]
\begin{center}
\includegraphics[clip,width=0.9\columnwidth]{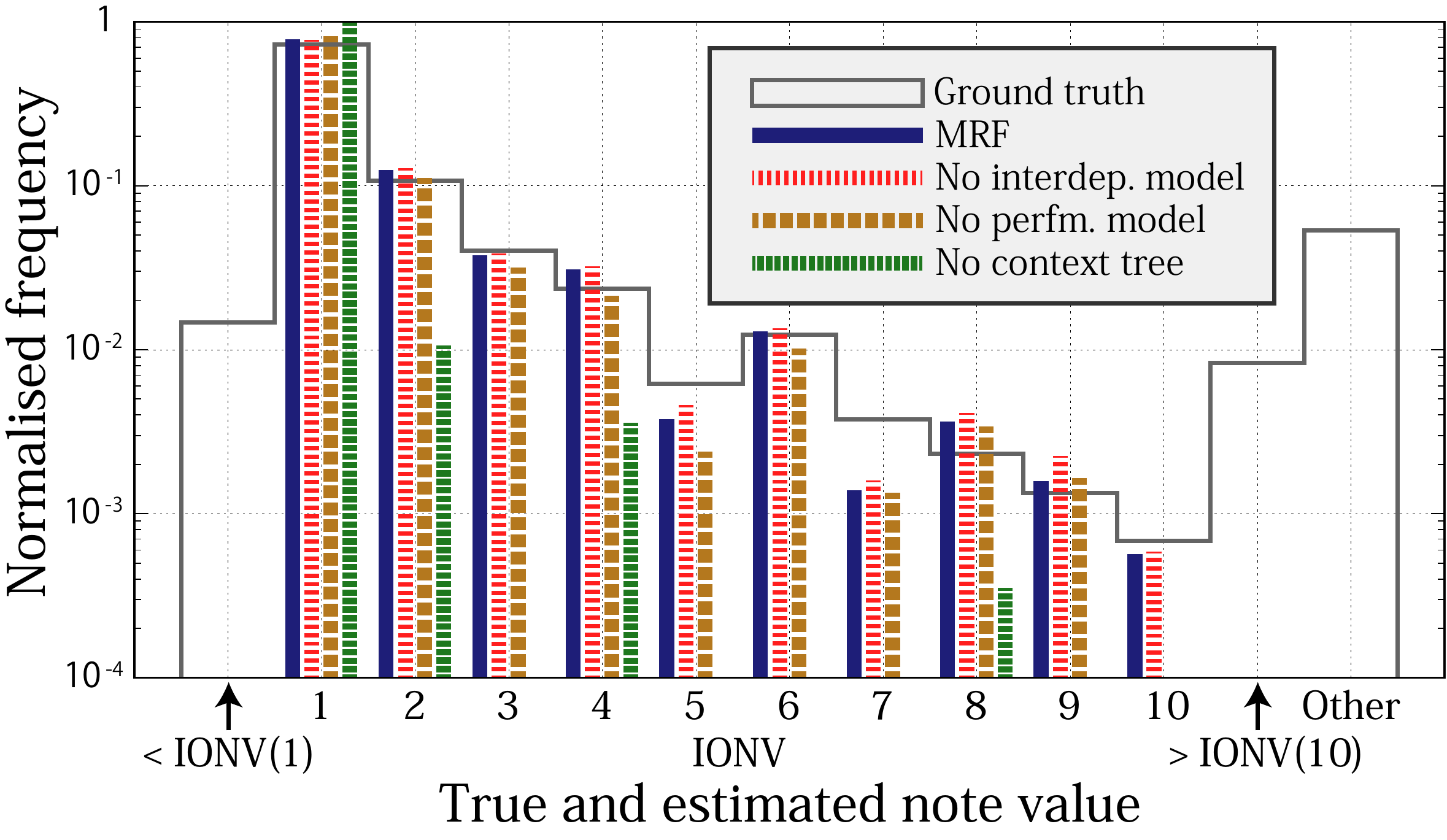}
\end{center}
\vspace{-3mm}
\caption{Distributions of true and estimated note values relative to IONVs.}
\label{fig:ErrorAnalysis}
\end{figure}
To examine the effect of the component models, let us look at the distribution of the estimated note values in the space of IONVs (Fig.~\ref{fig:ErrorAnalysis}).
Note that the distribution for the ground truth is essentially the same as that in Fig.~\ref{fig:DistrIONVClass}(b) but slightly different because the data is different and the onset clusters here are defined with the result of onset rhythm transcription by the metrical HMM.

Firstly, the model without a context tree assigns the first IONV to note values with a high probability ($>98\%$), indicating that estimated results by the model are almost the same as for the one-voice representation in Fig.~\ref{fig:ExampleScore}(b).
This is consistent with the results in Fig.~\ref{fig:NotewiseScaleError} that this model tends to estimate note values shorter than the correct values.
Secondly, one can notice that the model without the performance model has a higher probability for the first IONV and smaller probabilities for most of the later IONVs compared with the full model.
This suggests that the performance model uses the information of actual durations to correct (or better approximate) the estimated note values more frequently to larger values, leading to decreased scale errors.
Finally, the proportion of errors corresponding to note values that are larger than ${\rm IONV}(10)$ is about 0.8\%, indicating that the effect of enlarging the search space of note values by including higher IONVs is limited.

\subsubsection{Influence of the context-tree size and weights}

\begin{figure}[t]
\begin{center}
\includegraphics[clip,width=0.97\columnwidth]{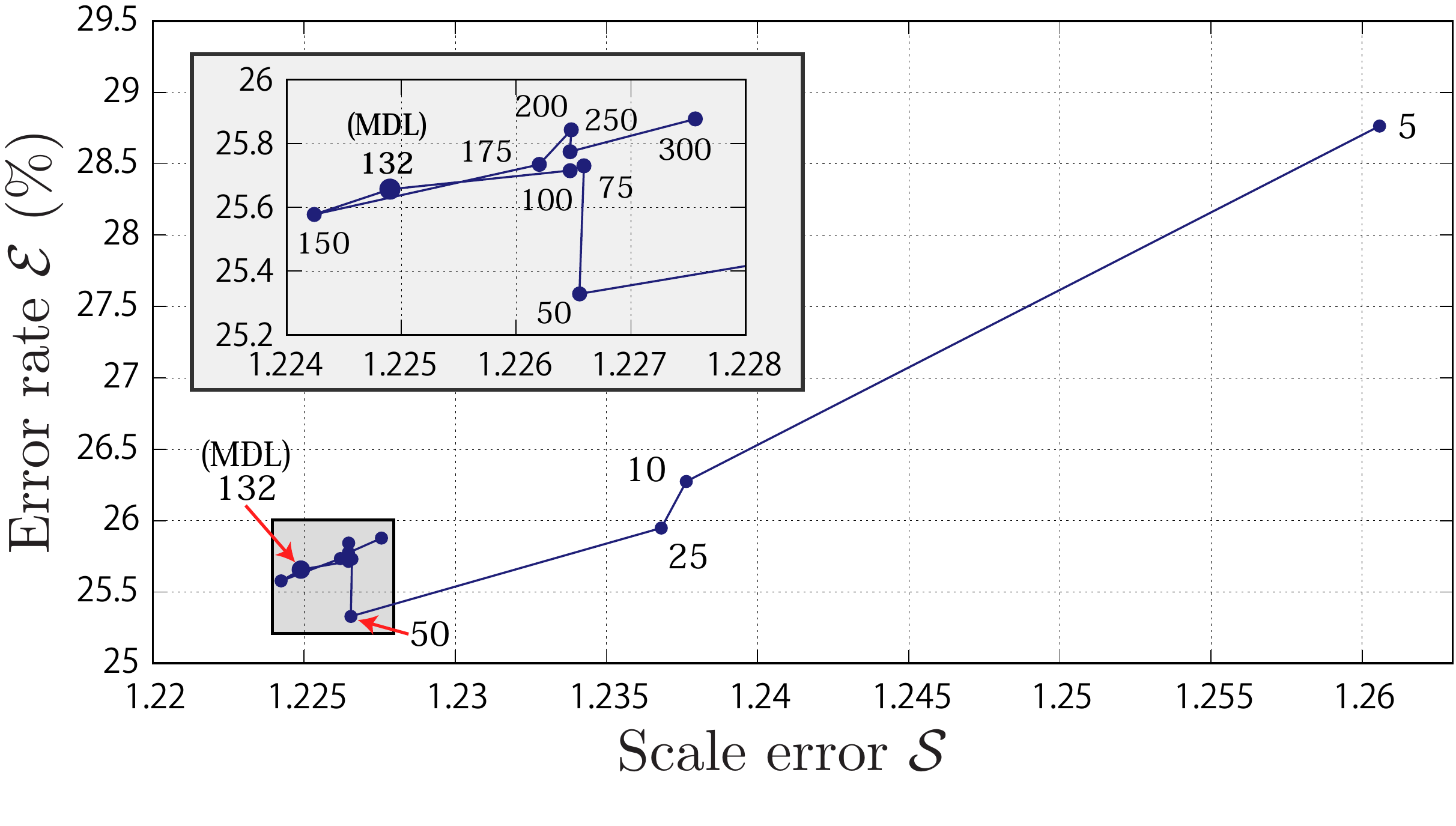}
\end{center}
\vspace{-5mm}
\caption{Average error rates and scale errors for various sizes of the context tree. The figure close to each point indicates the number of leaves. 132 is the optimal number predicted by the MDL criterion. All data points have statistical errors of order 1\% for error rate and order 0.01 for scale error.}
\label{fig:ERVarTreeSize}
\vspace{-3mm}
\end{figure}
Fig.~\ref{fig:ERVarTreeSize} shows the average error rates and scale errors for various sizes of the context tree.
The case with only one leaf (not shown in the figure) is the same as the case without a context tree explained above.
The errors rapidly decreased as the tree size increased for small numbers of leaves and but changed only slightly above 50 leaves.
There was a gap between the error rates for the cases with 50 and 75 leaves, which we confirmed is caused by a discontinuity of results for 52 and 53 leaves.
We have not succeeded in finding a good explanation for this gap.
Far above the predicted value (132 leaves) by the MDL criterion, the errors tended to increase slightly, confirming that it is close to the optimal choice.

\begin{figure}[t]
\begin{center}
\includegraphics[clip,width=0.97\columnwidth]{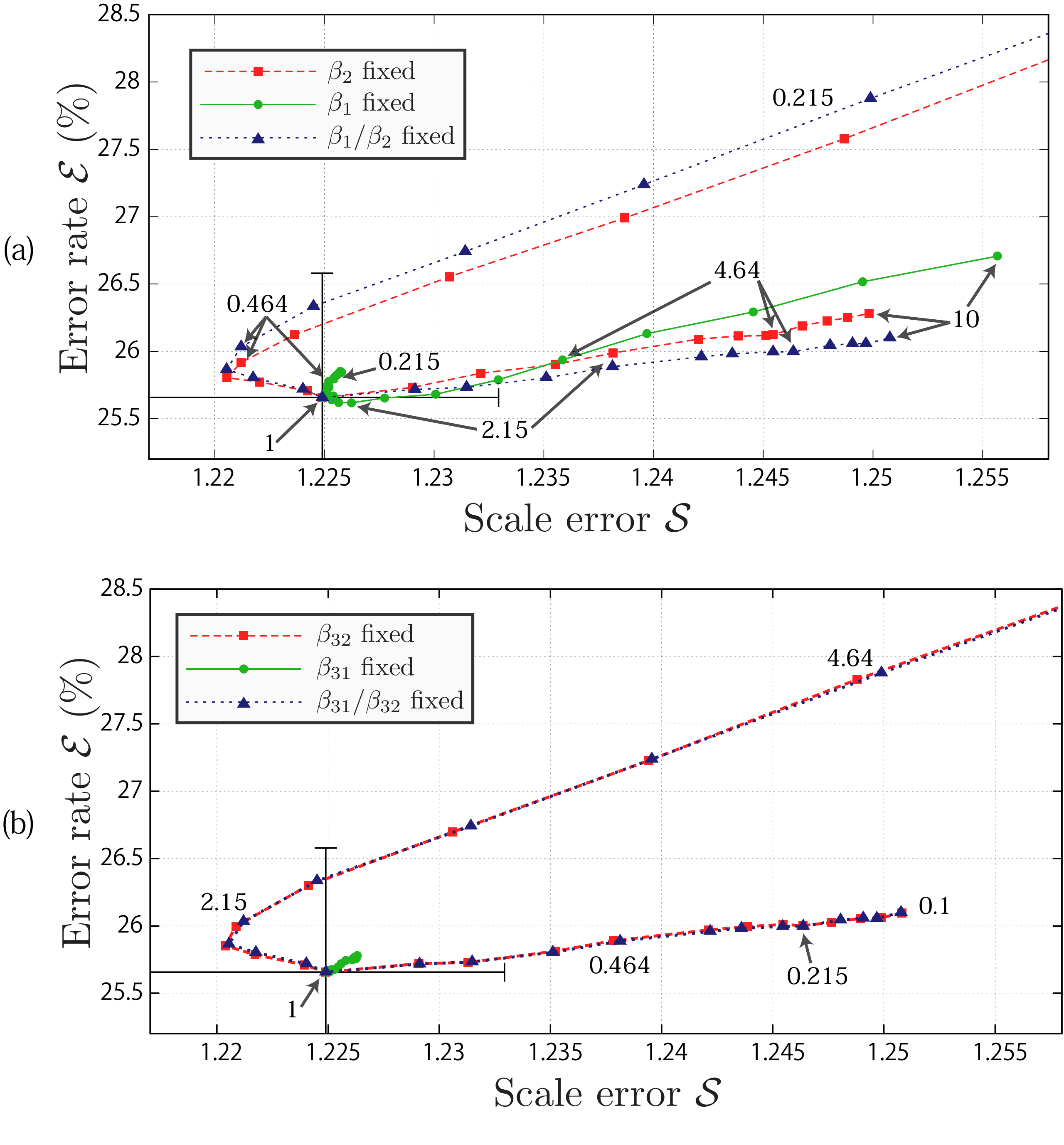}
\end{center}
\vspace{-4mm}
\caption{Average error rates and scale errors with (a) varying $\beta_1$ and $\beta_2$ and (b) varying $\beta_{31}$ and $\beta_{32}$. The $\beta$\,s are scaled in logarithmically equally spaced scaling factors, which are partly indicated by numbers, and the centre values (indicated by `$1$') are given in Eqs.~(\ref{eq:UsedBeta12}) and (\ref{eq:UsedBeta3132}). All data points have statistical errors of order 1\% for error rate and order 0.01 for scale error.}
\label{fig:InfluenceOfBeta}
\vspace{2mm}
\end{figure}
Fig.~\ref{fig:InfluenceOfBeta} shows the average error rate and scale error when varying the weights from the values in Eqs.~(\ref{eq:UsedBeta12}) and (\ref{eq:UsedBeta3132}).
The context tree had 132 leaves.
First, variations by increasing and decreasing the weights by 50\% are within $1\sigma$ statistical significance, showing that the error rates are not very sensitive to these parameters.
Second, the values $\hat{\beta}_1$ and $\hat{\beta}_2$, which were optimised based on ML using the score data, are found to be optimal with respect to the error rate.
Finally, the similar shapes of the curves when fixing $\beta_1/\beta_2$ and fixing $\beta_{31}/\beta_{32}$ show that their relative values influences the results more than their absolute values in the examined region.
The results together with the large-variance nature of the distributions of durations in Fig.~\ref{fig:Distribution} suggest that it is likely that more elaborate fitting functions for the performance model would not improve the results significantly and also that the sub-optimisation procedure for $\beta$\,s described in Sec.~\ref{sec:Optimisation} did not deteriorate the results much.

\subsubsection{Influence of the parameters of the performance model}
\label{sec:InfluenceOfPerfmModel}

\begin{table}[t]
\begin{center}
{\tabcolsep = 3pt
\begin{tabular}{cccc}\toprule
Key-holding $g$ & Damper-lifting $\bar{g}$ & Error rate ${\cal E}$ (\%) & Scale error ${\cal S}$\\
\midrule
Best fit & Best fit & $25.66$ & $1.225$ \\
Best fit & Trial 1  & $25.67$ & $1.225$ \\
Best fit & Trial 2  & $25.67$ & $1.225$ \\
Trial 1 & Best fit  & $25.97$ & $1.225$ \\
Trial 1 & Trial 1   & $25.98$ & $1.225$ \\
Trial 1 & Trial 2   & $25.97$ & $1.225$ \\
Trial 2 & Best fit  & $25.46$ & $1.225$ \\
Trial 2 & Trial 1   & $25.46$ & $1.225$ \\
Trial 2 & Trial 2   & $25.46$ & $1.225$ \\
\bottomrule
\end{tabular}
}
\end{center}
\caption{Average error rates and scale errors for different distributions for the performance model. The best fit and trial distributions are shown in Fig.~\ref{fig:VarPerfmModel}.}
\label{tab:VarPerfmModel}
\vspace{-6mm}
\end{table}
To examine the influence of the parameter values of the performance model in Eqs.~(\ref{eq:g}) and (\ref{eq:gbar}), we run the proposed model for each of three distributions shown in Figs.~\ref{fig:VarPerfmModel}(a) and \ref{fig:VarPerfmModel}(b).
The other parameters were set to the optimal values and the size of the context tree was $132$.
Results in Table \ref{tab:VarPerfmModel} show that despite the differences among distributions, the average scale error was almost constant and the variation of the average error rate is also smaller than the standard error.
More precisely, the influence of the choice of parameters for $\bar{g}$ is negligible, which can be explained by the small value of $\beta_{32}$.
This confirms that the influence of the performance model is small and there is little effect of overfitting in using the test data for learning.

\subsubsection{Example Result}

\begin{figure}[t]
\begin{center}
\includegraphics[clip,width=1.\columnwidth]{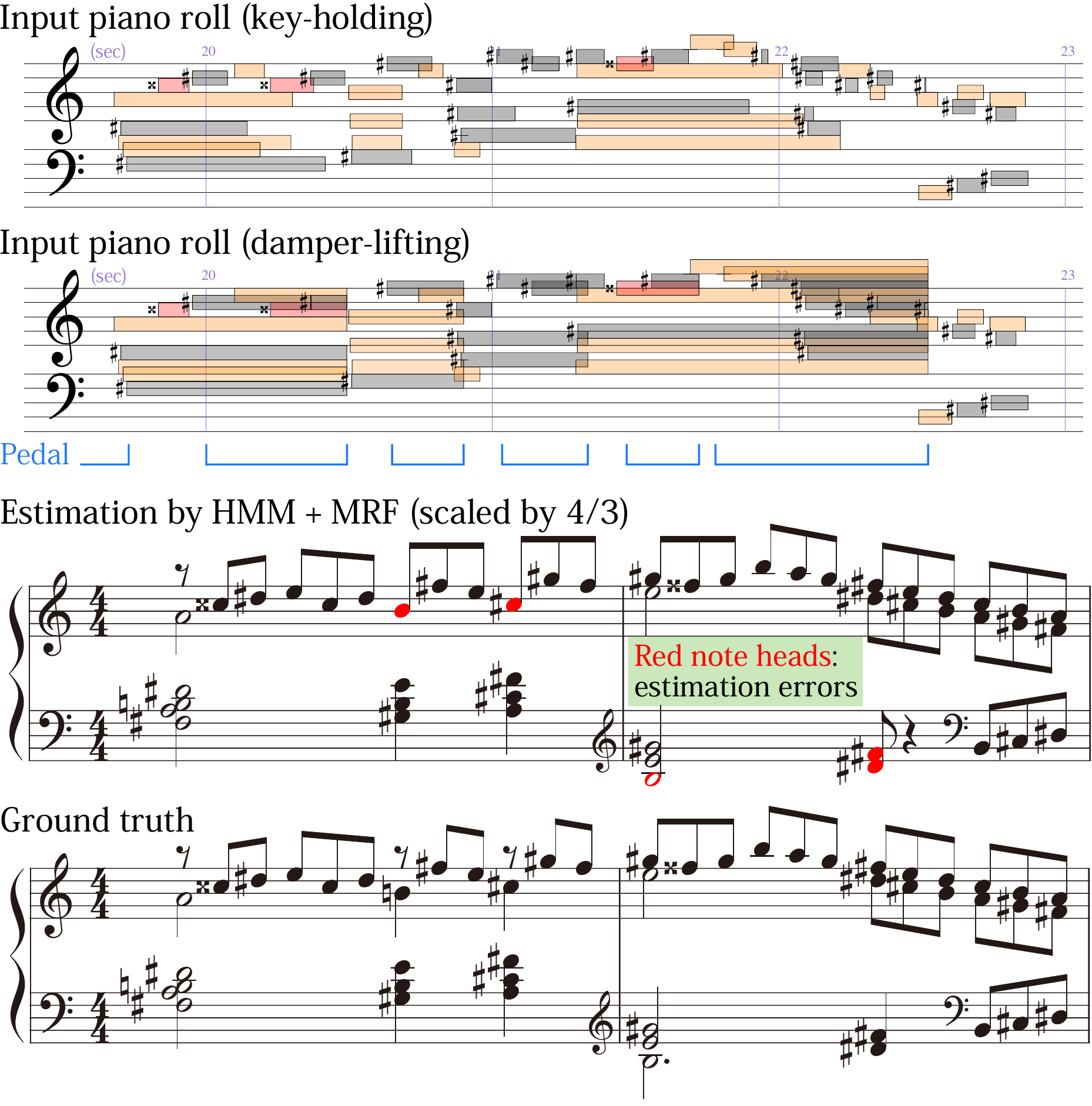}
\end{center}
\vspace{-5mm}
\caption{Example result of rhythm transcription by the metrical HMM and the proposed MRF model (Beethoven: Waldstein sonata 1st mov.). Voice, staff and time signature are added manually to the estimated result for the purpose of this illustration.}
\label{fig:ExResult}
\vspace{-3mm}
\end{figure}
Let us discuss an example\footnote{Sound files are available at the accompanying web page \cite{Webpage}.} in Fig.~\ref{fig:ExResult}, which has a typical texture of piano music with the left-hand part having harmonising chords and the right-hand part having melodic notes, both of which have multiple voices inside.
By comparing the performed durations to the score, we can see that overall the damper-lifting durations are closer to the score-indicated durations for the left-hand notes and the key-holding durations are closer for the right-hand notes.
This is because pianists tend to lift the pedal when harmonising chords change.
This example shows that the two types of durations provide complementary information and one should not rely on one of them.
On the other hand, for most notes, the offset score time matches to the onset score time of a succeeding note with a close pitch, which is what our context model describes.

The result by the MRF model shows that the model uses the score and performance models complementarily to find the optimal estimation.
The correctly estimated half notes (as ${\rm IONV}(6)$), A4 in the first bar and E5 in the second bar, have a close pitch in the next onset cluster and the incorrect estimates as ${\rm IONV}(1)$ are avoided by using the duration (and perhaps because of the existence of very close pitches at the sixth next onset clusters).
On the other hand, the quarter-note F\#4 and D\#4 in the left-hand part in the second bar could not be correctly estimated probably because the voice makes a big leap here, closer notes in the right-hand part succeed them and the key-holding durations are short.

\section{Conclusion and Discussion}

We discussed note value recognition of polyphonic piano music based on an MRF model combining the score model and the performance model.
As suggested in the discussion in Sec.~\ref{sec:Observation} and confirmed by evaluation results, performed durations can deviate greatly from the score-indicated lengths and thus the performance model aline has little predictive ability.
The construction of the score model is then the key to solve the problem.
We formulated a context-tree model that can learn highly predictive distributions of note values from data, using onset score times and the pitch context.
It was demonstrated that this score model brings significant improvements on the recognition accuracy.

Refinement of the score model is possible in a number of ways.
Using more features for the context-tree model could improve the results.
Using other feature-based model learning schemes such as deep neural networks are similarly possible.
The refinement and extension of the search space for note values is another issue since the set of the first ten IONVs used in this study loses a certain proportion of solutions.
The result that the context-tree model learned to capture the voice structure suggests that building a model with explicit voice structure is also interesting for creating generative models to reduce reliance on arbitrarily chosen features.

Remaining issues to obtain musical scores in a fully automatic way include the assignment of voice and staff to the transcribed notes.
Voice separation methods and staff estimation methods exist (e.g.\ \cite{Cambouropoulos2008,McLeod2016,HandSeparation}) and the information of transcribed note values can be useful to identify chordal notes within each voice.
Another issue is the recognition of time signature.
Using multiple metrical HMMs learned with score data for each metres is one possibility and we could also apply other metre detection methods (e.g.\ \cite{Haas2016}) to the transcribed result.

To apply this work, the construction of a complete polyphonic music transcription system from audio signals to musical scores is attractive.
The framework developed in this study can be combined with existing multi-pitch analysers \cite{Vincent2010,OHanlon2014,Yoshii2015,Sigtia2016} for this purpose.
It is worth mentioning that the performance model should be trained on piano rolls obtained with these methods since the distribution of durations would differ from that of recorded MIDI signals.
Extension of the model to correct audio transcription errors such as note insertions and deletions would also be of great importance.

\section*{Acknowledgement}
We are grateful to David Temperley for providing source code for the Melisma Analyzer.
E.~Nakamura would like to thank Shinji Takaki for useful discussions about context-tree clustering.
This work is in part supported by JSPS KAKENHI Nos.\ 24220006, 26280089, 26700020, 15K16054, 16H01744 and 16J05486, JST ACCEL No.\ JPMJAC1602, and the long-term overseas research fund by the Telecommunications Advancement Foundation.



%


\begin{IEEEbiographynophoto}{Eita Nakamura}
He received a Ph.D.\ degree in physics from the University of Tokyo in 2012. After having been a post-doctoral researcher at the National Institute of Informatics, Meiji University and Kyoto University, he is currently a JSPS Research Fellow in the Speech and Audio Processing Group at Kyoto University. His research interests include music modelling and analysis, music information processing and statistical machine learning.
\end{IEEEbiographynophoto}

\begin{IEEEbiography}[{\includegraphics[width=1in,height=1.25in,clip,keepaspectratio]{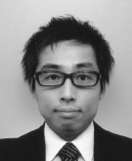}}]{Kazuyoshi Yoshii}
He received the Ph.D. degree in informatics from Kyoto University, Japan, in 2008. He is currently a Senior Lecturer at Kyoto University. His research interests include music signal processing and machine learning. He is a Member of the Information Processing Society of Japan and Institute of Electronics, Information, and Communication Engineers.
\end{IEEEbiography}

\begin{IEEEbiography}[{\includegraphics[width=1in,height=1.25in,clip,keepaspectratio]{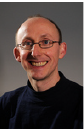}}]{Simon Dixon}
He is a Reader (Assoc. Prof.), Director of Graduate Studies and Deputy Director of the Centre for Digital Music at Queen Mary University of London. He has a Ph.D. in Computer Science (Sydney) and LMusA diploma in Classical Guitar.  His research interests include high-level music signal analysis, computational modelling of musical knowledge, and the study of musical performance. Particular areas of focus include automatic music transcription, beat tracking, audio alignment and analysis of intonation and temperament. He was President (2014-15) of the International Society for Music Information Retrieval (ISMIR), is founding Editor of the Transactions of ISMIR, and member of the Editorial Board of the Journal of New Music Research (since 2011), and has published over 160 refereed papers in the area of music informatics.
\end{IEEEbiography}

\end{document}